\documentclass[10pt,twocolumn,letterpaper]{article}

\usepackage[pagenumbers]{cvpr} 

\usepackage{graphicx}
\usepackage{amsmath}
\usepackage{epsfig}
\usepackage{amssymb}
\usepackage{enumitem}
\usepackage[normalem]{ulem}
\usepackage[ruled]{algorithm2e}
\usepackage{multirow}
\usepackage{array}
\usepackage{ctable} 
\usepackage{makecell}
\usepackage[capposition=bottom]{floatrow}
\usepackage{comment}
\DeclareMathAlphabet{\mathcal}{OMS}{cmsy}{m}{n}
\usepackage{textcomp}
\usepackage{dblfloatfix}
\usepackage{caption}
\usepackage{subcaption}
\usepackage{booktabs}
\usepackage{mathtools}
\usepackage{textpos}
\usepackage{minitoc}
\usepackage{booktabs}
\usepackage[toc,page,header]{appendix}
\usepackage{color}
\usepackage{tabularx}
\usepackage{bbding}
\usepackage{pifont}

\newcommand{\namecvpr}{\textsc{ObjectFolder}\xspace}
\newcommand{\namereal}{\textsc{ObjectFolder Real}\xspace}
\newcommand{\name}{\textsc{ObjectFolder Benchmark}\xspace}

\definecolor{nicegreen}{rgb}{0.1, 0.6, 0.2}

\definecolor{citecolor}{RGB}{34,139,34}
\usepackage[pagebackref=true,breaklinks=true,letterpaper=true,colorlinks,
citecolor=citecolor,bookmarks=false]{hyperref}

\usepackage[capitalize]{cleveref}
\crefname{section}{Sec.}{Secs.}
\Crefname{section}{Section}{Sections}
\Crefname{table}{Table}{Tables}
\crefname{table}{Tab.}{Tabs.}

\def\cvprPaperID{3951} 
\def\confName{CVPR}
\def\confYear{2023}

\begin{document}


\title{The \name: \\ Multisensory Learning with \emph{Neural} and \emph{Real} Objects}
\author{Ruohan Gao\thanks{indicates equal contribution.} \hspace{5mm} Yiming Dou\footnotemark[1] \thanks{Yiming is affiliated with Shanghai Jiao Tong University. The work was done when he was visiting Stanford University as a summer intern.}  \hspace{5mm}  Hao Li\footnotemark[1] \hspace{5mm} Tanmay Agarwal \hspace{5mm} Jeannette Bohg \hspace{5mm} Yunzhu Li \\ Li Fei-Fei \hspace{10mm} Jiajun Wu\\
Stanford Univeristy\\
}
\maketitle

\begin{abstract}
We introduce the \textsc{ObjectFolder Benchmark}, a benchmark suite of 10 tasks for multisensory object-centric learning, centered around object recognition, reconstruction, and manipulation with sight, sound, and touch. We also introduce the \namereal dataset, including the multisensory measurements for 100 real-world household objects, building upon a newly designed pipeline for collecting the 3D meshes, videos, impact sounds, and tactile readings of real-world objects. We conduct systematic benchmarking on both the 1,000 multisensory neural objects from \namecvpr, and the real multisensory data from \namereal. Our results demonstrate the importance of multisensory perception and reveal the respective roles of vision, audio, and touch for different object-centric learning tasks. 
By publicly releasing our dataset and benchmark suite, we hope to catalyze and enable new research in multisensory object-centric learning in computer vision, robotics, and beyond. Project page: \url{https://objectfolder.stanford.edu}
\end{abstract}


 \doparttoc %
\faketableofcontents %

\part{} %
%

\vspace{-0.45in}
\section{Introduction}\label{sec:intro}

Computer vision systems today excel at recognizing objects in 2D images thanks to many image datasets~\cite{deng2009imagenet,lin2014microsoft,barbu2019objectnet,kuznetsova2020open}.  There is also a growing interest in modeling an object's shape and appearance in 3D, with various benchmarks and tasks introduced~\cite{chang2015shapenet,pix3d,reizenstein2021common,greff2022kubric,mescheder2019occupancy,mildenhall2020nerf}. Despite the exciting progress, these studies primarily focus on the visual recognition of objects. At the same time, our everyday activities often involve multiple sensory modalities. Objects exist not just as \emph{visual} entities, but they also make sounds and can be touched during interactions. The different sensory modes of an object all share the same underlying object intrinsics---its 3D shape, material property, and texture. Modeling the complete multisensory profile of objects is of great importance for many applications beyond computer vision, such as robotics, graphics, and virtual and augmented reality.

Some recent attempts have been made to combine multiple sensory modalities to complement vision for various tasks~\cite{smith20203d,smith2021active,dar,zhang2017shape,calandra2018more,li2019connecting,Suresh22icra,yang2022touch}. These tasks are often studied in tailored settings and evaluated on different datasets. As an attempt to develop assets generally applicable to diverse tasks, the \namecvpr dataset~\cite{gao2021ObjectFolder,gao2022ObjectFolderV2} has been introduced and includes 1,000 neural objects with their visual, acoustic, and tactile properties. \namecvpr however has two fundamental limitations. First, no real objects are included; all multisensory data are obtained through simulation with no simulation-to-real (sim2real) calibration. Second, only a few tasks were presented to demonstrate the usefulness of the dataset and to establish the possibility of conducting sim2real transfer with the neural objects. 

\begin{figure*}
    \center
    \includegraphics[scale=0.83]{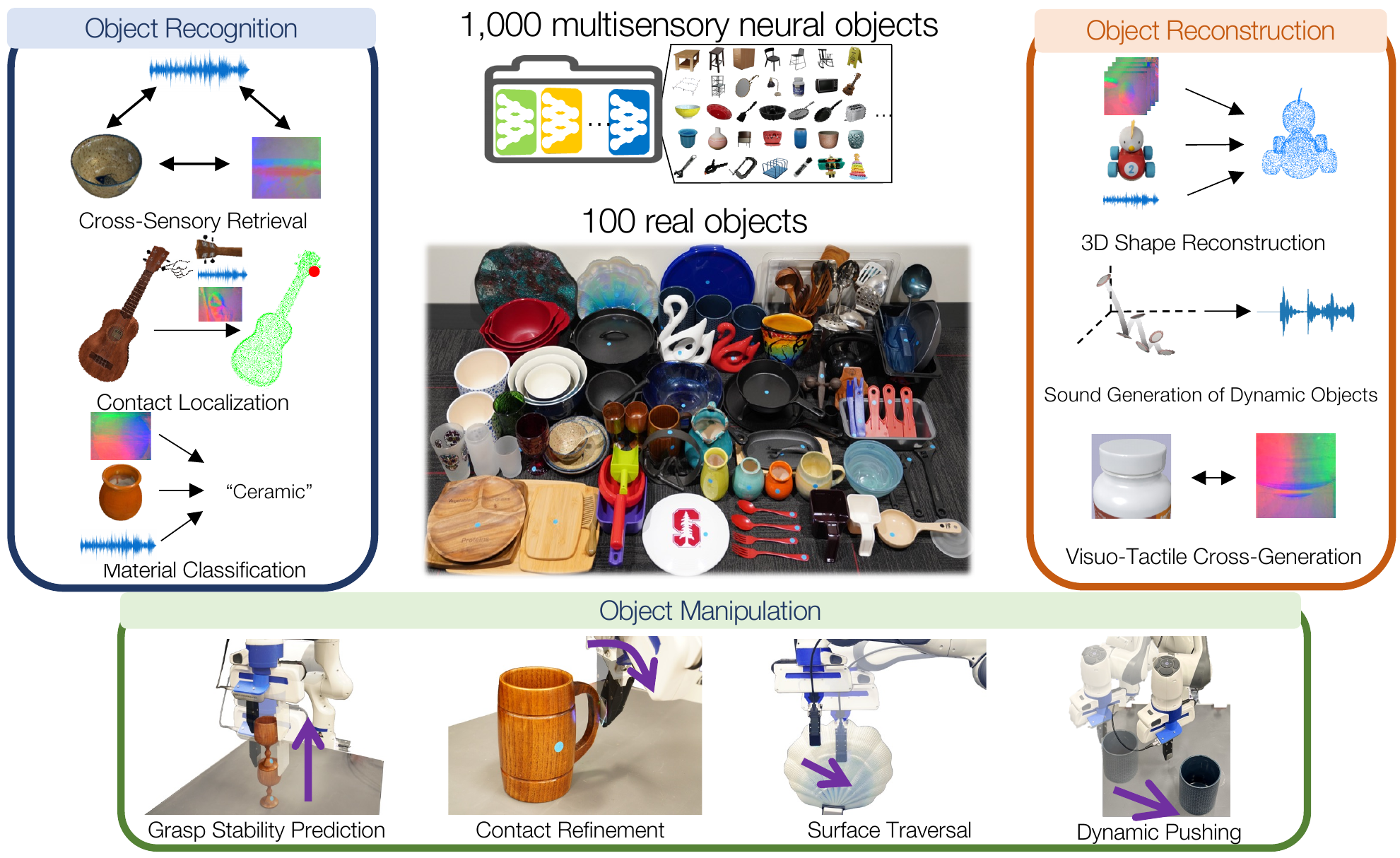}
    \vspace{-0.05in}
    \caption{The \name suite consists of 10 benchmark tasks for multisensory object-centric learning, centered around object recognition, reconstruction, and manipulation. Complementing the 1,000 multisensory neural objects from \namecvpr~\cite{gao2022ObjectFolderV2}, we also introduce \namereal, which contains real multisensory data collected from 100 real-world objects, including their 3D meshes, video recordings, impact sounds, and tactile readings.}
    \label{fig:concept}
    \vspace{-0.05in}
\end{figure*}

Consequently, we need a multisensory dataset of real objects and a robust benchmark suite for multisensory object-centric learning. To this end, we present the \namereal dataset and the \name suite, as shown in Fig.~\ref{fig:concept}. 

The \namereal dataset contains multisensory data collected from 100 real-world household objects. We design a data collection pipeline for each modality: for vision, we scan the 3D meshes of objects in a dark room and record HD videos of each object rotating in a lightbox; for audio, we build a professional anechoic chamber with a tailored object platform and then collect impact sounds by striking the objects at different surface locations with an impact hammer; for touch, we equip a Franka Emika Panda robot arm with a GelSight robotic finger~\cite{yuan2017gelsight,dong2017improved} and collect tactile readings at the exact surface locations where impact sounds are collected.

The \name suite consists of 10 benchmark tasks for multisensory object-centric learning, centered around object recognition, reconstruction, and manipulation. The three recognition tasks are cross-sensory retrieval, contact localization, and material classification; the three reconstruction tasks are 3D shape reconstruction, sound generation of dynamic objects, and visuo-tactile cross-generation; and the four manipulation tasks are grasp stability prediction, contact refinement, surface traversal, and dynamic pushing. We standardize the task setting for each task and present baseline approaches and results.

Experiments on both neural and real objects demonstrate the distinct value of sight, sound, and touch in different tasks. For recognition, vision and audio tend to be more reliable compared to touch, where the contained information is too local to recognize. For reconstruction, we observe that fusing multiple sensory modalities achieve the best results, and it is possible to hallucinate one modality from the other. This agrees with the notion of degeneracy in cognitive studies~\cite{smith2005development}, which creates redundancy such that our sensory system functions even with the loss of one component. For manipulation, vision usually provides global positional information of the objects and the robot, but often suffers from occlusion. Touch, often as a good complement to vision, is especially useful to capture the accurate local geometry of the contact point.

We will open-source all code and data for \namereal and \name to facilitate research in multisensory object-centric learning.

\section{Related Work}\label{sec:related}

\paragraph{Object Datasets.} A large body of work in computer vision focuses on recognizing objects in 2D images~\cite{krizhevsky2017imagenet,girshick2014rich,he2016deep,he2017mask}. This progress is enabled by a series of image datasets such as ImageNet~\cite{deng2009imagenet}, MS COCO~\cite{lin2014microsoft}, ObjectNet~\cite{barbu2019objectnet}, and OpenImages~\cite{kuznetsova2020open}. In 3D vision, datasets like ModelNet~\cite{wu20153d} and ShapeNet~\cite{chang2015shapenet} focus on modeling the geometry of objects but without realistic visual textures. Recently, with the popularity of neural rendering approaches~\cite{neff2021donerf,sitzmann2019scene}, a series of 3D datasets are introduced with both realistic shape and appearance, such as CO3D~\cite{reizenstein2021common},
Google Scanned Objects~\cite{downs2022google}, and
ABO~\cite{collins2021abo}. Unlike all datasets above that focus only on the visual modality, we also model the acoustic and tactile modalities of objects.

Our work is most related to \namecvpr~\cite{gao2021ObjectFolder,gao2022ObjectFolderV2}, a dataset of 1,000 neural objects with visual, acoustic, and tactile sensory data. While their multisensory data are obtained purely from simulation, we introduce the \namereal dataset that contains real multisensory data collected from real-world household objects.

\vspace{-0.15in}
\paragraph{Capturing Multisensory Data from Real-World Objects.} Limited prior work has attempted to capture multisensory data from the real world. Earlier work models the multisensory physical behavior of 3D objects~\cite{pai2001scanning} for virtual object interaction and animations. To our best knowledge, there is no large prior dataset of real object impact sounds. Datasets of real tactile data are often collected for a particular task such as robotic grasping~\cite{calandra2017feeling,calandra2018more}, cross-sensory prediction~\cite{li2019connecting}, or from unconstrained in-the-wild settings~\cite{yang2022touch}. Our \namereal dataset is the first dataset that contains all three modalities with rich annotations to facilitate multisensory learning research with real object data.

\begin{figure*}[t!]
    \centering
    \begin{subfigure}[b]{.33\linewidth}
        \centering
        \vspace{-0.1in}
        \includegraphics[scale=0.755]{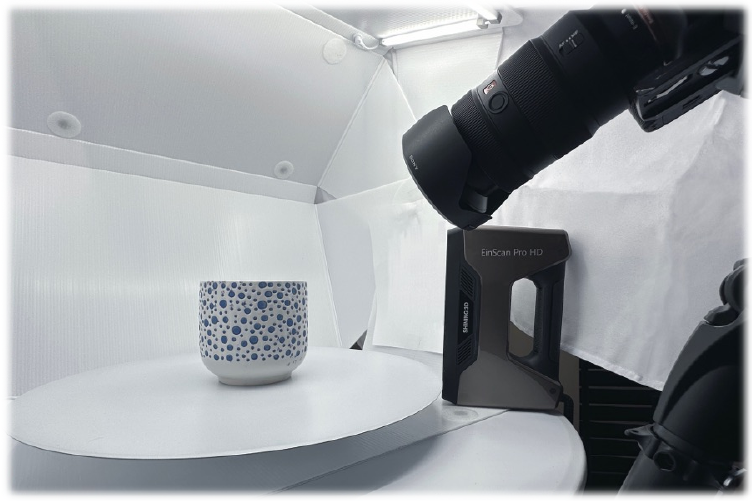}
        \caption{Visual data collection}
        \label{fig:setup_vision}
    \end{subfigure}
    \hfill
    \begin{subfigure}[b]{.33\linewidth}
        \centering
        \includegraphics[scale=0.755]{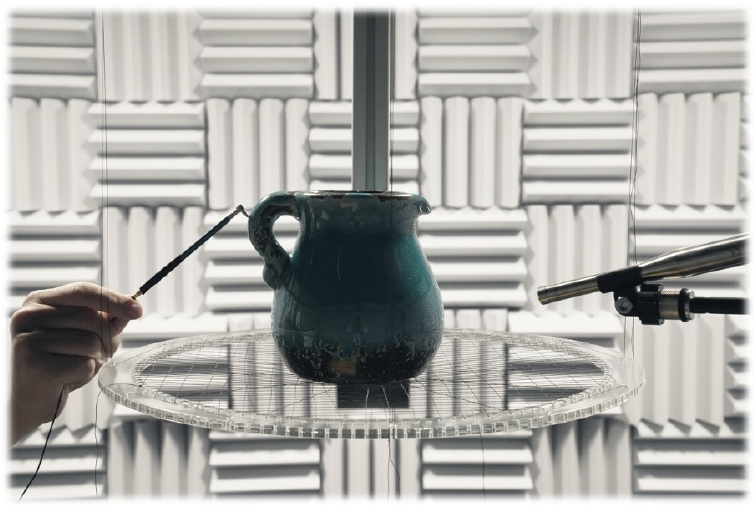}
        \caption{Acoustic data collection}
        \label{fig:setup_audio}
    \end{subfigure}
    \hfill
    \begin{subfigure}[b]{.33\linewidth}
        \centering
        \includegraphics[scale=0.755]{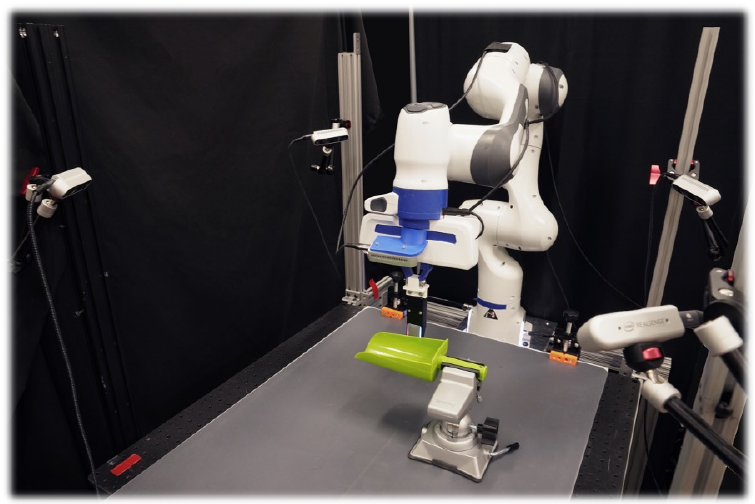}
        \caption{Tactile data collection}
        \label{fig:setup_touch}
    \end{subfigure}
    \vspace{-0.25in}
    \caption{Illustration of our multisensory data collection pipeline for the \namereal dataset. We design a tailored hardware solution for each sensory modality to collect high-fidelity visual, acoustic, and tactile data for 100 real household objects.}
    \vspace{-0.05in}
    \label{fig:data_collection}
\end{figure*}

\vspace{-0.15in}
\paragraph{Multisensory Object-Centric Learning.} Recent work uses audio and touch in conjunction with vision for a series of new tasks, including visuo-tactile 3D reconstruction~\cite{smith20203d,smith2021active,Suresh22icra,gao2022ObjectFolderV2}, cross-sensory retrieval~\cite{gao2021ObjectFolder,dar}, cross-modal generation~\cite{zhang2017shape,li2019connecting,lee2019touching}, contact localization~\cite{gao2022ObjectFolderV2,luo2015localizing}, robotic manipulation~\cite{calandra2018more,calandra2017feeling,lee2019making,li2022seehearfeel}, and audio-visual learning from videos~\cite{gao2018objectSounds,owens2018audio,gao2019co,arandjelovic2017look,zhao2018sound,chen2022visual,chen2022sound}. While they only focus on a single task of interest in tailored settings, each with a different set of objects, we present a standard benchmark suite of 10 tasks based on 1,000 neural objects from \namecvpr and 100 real objects from \namereal for multisensory object-centric learning.
\vspace{-0.05in}
\section{\namereal}\label{sec:dataset}
\vspace{-0.05in}

The \namecvpr dataset~\cite{gao2022ObjectFolderV2} contains 1,000 multisensory neural objects, each represented by an \emph{Object File}, a compact neural network that encodes the object's intrinsic visual, acoustic, and tactile sensory data. Querying it with extrinsic parameters (\eg, camera viewpoint and lighting conditions for vision, impact location and strength for audio, contact location and gel deformation for touch), we can obtain the corresponding sensory signal at a particular location or condition.

Though learning with these virtualized objects with simulated multisensory data is exciting, it is necessary to have a benchmark dataset of multisensory data collected from real objects to quantify the difference between simulation and reality. Having a well-calibrated dataset of real multisensory measurements allows researchers to benchmark different object-centric learning tasks on real object data without having the need to actually acquire these objects. For tasks in our benchmark suite in Sec.~\ref{sec:benchmark}, we show results on both the neural objects from \namecvpr and the real objects from \namereal when applicable.

Collecting real multisensory data densely
from real objects is very challenging, requiring careful hardware design and tailored solutions for each sensory modality by taking into account the physical
constraints (e.g., robot joint limit, kinematic constraints) in the capture system. Next, we introduce how we collect the visual (Sec.~\ref{sec:visual_data}), acoustic (Sec.~\ref{sec:acoustic_data}), and tactile (Sec.~\ref{sec:tactile_data}) data for the 100 real objects shown in Fig.~\ref{fig:concept}. Please also visit our project page for interactive demos to visualize the captured multisensory data.

\subsection{Visual Data Collection}\label{sec:visual_data}

We use an EinScan Pro HD 2020 handheld 3D Scanner\footnote{\url{https://www.einscan.com}} to scan a high-quality 3D mesh and the corresponding color texture for each object. The scanner captures highly accurate 3D features by projecting a visible light array on the object and records the texture through an attached camera. The minimum distance between two points in the scanned point cloud is $0.2~mm$, enabling fine-grained details of the object's surface to be retained in the scanned mesh. For each object, we provide three versions of its mesh with different resolutions: $16\text{K}$ triangles, $64\text{K}$ triangles, and Full resolution (the highest number of triangles possible to achieve with the scanner). Additionally, we record an HD video of each object rotating in a lightbox with a professional camera to capture its visual appearance, as shown in Fig.~\ref{fig:setup_vision}.

\subsection{Acoustic Data Collection}\label{sec:acoustic_data}

We use a professional recording studio with its walls treated with acoustic melamine anechoic foam panels and the ceiling covered by absorbing acoustic ceiling tiles, as shown in Fig.~\ref{fig:setup_audio}. The specific setup used to collect audio data varies with the object's weight and size. Most objects are placed on a circular platform made with thin strings, which minimally affects the object's vibration pattern when struck. Light objects are hung with a thin string and hit while suspended in the air. Heavy objects are placed on top of an anechoic foam panel to collect their impact sounds. 

For each object, we select 30--50 points based on its scale following two criteria. First, the points should roughly cover the whole surface of the object and reveal its shape; Second, we prioritize points with specific local geometry or texture features, such as the rim/handle of a cup. For each selected point, we collect a 5-second audio clip of striking it along its normal direction with a PCB\footnote{\url{https://www.pcb.com}} impact hammer (086C01). The impact hammer is equipped with a force transducer in its tip, providing ground-truth contact forces synchronized with the audio recorded by a PCB phantom-powered free-field microphone (376A32). It is made of hardened steel, which ensures that the impacts are sharp and short enough to excite the higher-frequency modes of each object. We also record the accompanying video with a RealSense RGBD camera along with each impact sound.

\subsection{Tactile Data Collection}\label{sec:tactile_data}

Fig.~\ref{fig:setup_touch} illustrates our setup for the tactile data collection. We equip a Franka Emika Panda robot arm with a GelSight touch sensor~\cite{yuan2017gelsight,dong2017improved} to automate the data collection process. GelSight sensors are vision-based tactile sensors that measure the texture and geometry of a contact surface with high spatial resolution through an elastomer and an embedded camera. We use the R1.5 GelSight tactile robot finger\footnote{\url{https://www.gelsight.com}}, which has a sensing area of $32 \times 24$ $mm^2$.

We mount a RealSense RGBD camera at each corner of the robot frame. After camera calibration, we use the RealSense ROS package to get a point cloud estimation of the target object. We also extract a point cloud from the scanned 3D mesh of the object. In order to align the two point clouds, we first manually select four roughly corresponding points on both point clouds to provide an initial registration. Next, we use the Iterative Closest Point (ICP)~\cite{besl1992method} algorithm for point cloud alignment. We add a manual adjustment step for cases where the ICP alignment is not accurate.

We collect tactile data at the same set of surface points where the impact sounds are collected for each object. For each point of interest, we provide the robot with the target position and orientation of the GelSight robot finger; we then use position control to automatically reach the target point following the normal direction of the target point. The robot finger stops when the tactile sensor cannot deform further. We collect a video of the tactile RGB images that record the gel deformation process. We also use an in-hand camera and a third-view camera to capture two videos of the contact process for each point.
\vspace{-0.05in}
\section{ObjectFolder Benchmark Suite}\label{sec:benchmark}
\vspace{-0.05in}

Our everyday activities involve the perception and manipulation of various objects. Modeling and understanding the multisensory signals of objects can potentially benefit many applications in computer vision, robotics, virtual reality, and augmented reality. The sensory streams of sight, sound, and touch all share the same underlying object intrinsics. During interactions, they often work together to reveal the object's category, 3D shape, texture, material, and physical properties.

Motivated by these observations, we introduce a suite of 10 benchmark tasks for multisensory object-centric learning, centered around \emph{object recognition} (Sec.~\ref{task:retrieval}, \ref{task:contact_localization}, and \ref{task:material}), \emph{object reconstruction} (Sec.~\ref{task:3dreconstruction}, \ref{task:sound_generation}, and \ref{task:visuo_tactile_generation}), and \emph{object manipulation} (Sec.~\ref{task:grasp_stability}, \ref{task:refinement}, \ref{task:surface_traversal}, and \ref{task:dynamic_pushing}), as shown in Fig.~\ref{fig:concept}. In the sections below, we first present the motivation for each task. Then, we standardize the task setting, define evaluation metrics, draw its connection to existing tasks, and develop baseline models leveraging state-of-the-art components from the literature. In the end, we show a teaser result for each task. \textbf{Please see Supp. for the complete results, baselines, and experimental setups.}

\vspace{-0.02in}
\subsection{Cross-Sensory Retrieval}\label{task:retrieval}
\vspace{-0.03in}

\paragraph{Motivation} 
When seeing a wine glass, we can mentally link how it looks to how it may sound when struck or feel when touched. For machine perception, cross-sensory retrieval also plays a crucial role in understanding the relationships between different sensory modalities. While existing cross-modal retrieval benchmarks and datasets~\cite{nuswide, peng2017overview, peng2018modalityspecific, Wikipedia, pascal_sentence} mainly focus on retrieval between images and text, we perform cross-sensory retrieval between objects' visual images, impact sounds, and tactile readings.

\vspace{-0.15in}
\paragraph{Task Definition.}
Cross-sensory retrieval requires the model to take one sensory modality as input and retrieve the corresponding data of another modality. For instance, given the sound of striking a mug, the ``audio2vision'' model needs to retrieve the corresponding image of the mug from a pool of images of hundreds of objects. In this benchmark, each sensory modality (vision, audio, touch) can be used as either input or output, leading to $9$ sub-tasks.

\begin{figure}
    \center
    \includegraphics[scale=0.29]{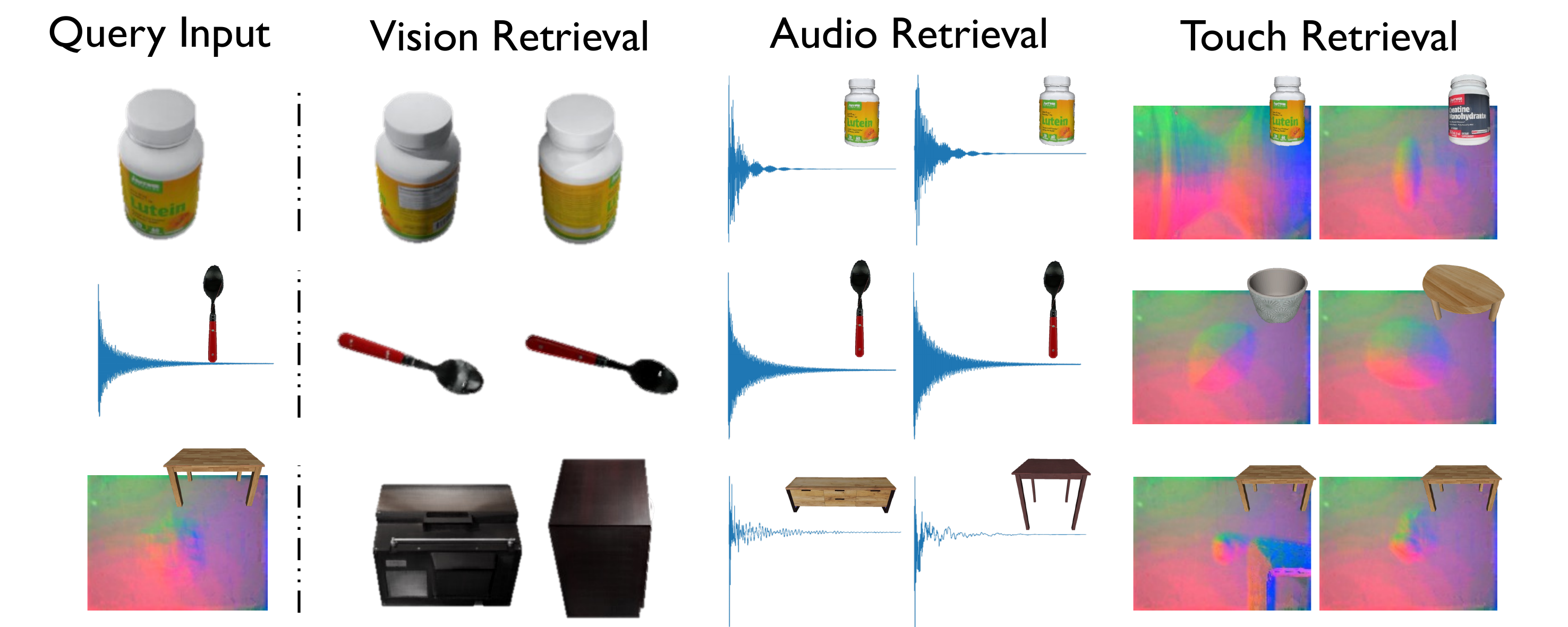}
    \vspace{-0.1in}
    \caption{Examples of the top-2 retrieved instances for each modality using DAR~\cite{dar}, the best-performing baseline. For audio and touch retrieval, we also show an image of the object.
    }
    \label{fig:retrieval_teaser_result}
\end{figure}

\vspace{-0.15in}
\paragraph{Evaluation Metrics and Baselines.}
We measure the mean Average Precision (mAP) score, a standard metric for evaluating retrieval. We adopt several state-of-the-art methods as the baselines: 1) Canonical Correlation Analysis (CCA)~\cite{cca}, 2) Partial Least Squares (PLSCA)~\cite{plsca}, 3) Deep Aligned Representations (DAR)~\cite{dar}, and 4) Deep Supervised Cross-Modal Retrieval (DSCMR)~\cite{dscmr}.  

\vspace{-0.15in}
\paragraph{Teaser Results.}
Fig.~\ref{fig:retrieval_teaser_result} shows examples of the top retrieved instances for DAR~\cite{dar}, the best-performing baseline. We can see that vision and audio tend to be more reliable for retrieval, while a single touch reading usually does not contain sufficient discriminative cues to identify an object.

\vspace{-0.05in}
\subsection{Contact Localization}\label{task:contact_localization}

\paragraph{Motivation.}
Localizing the contact point when interacting with an object is of great interest, especially for robot manipulation tasks. Each modality offers complementary cues: vision displays the global visual appearance of the contacting object; touch offers precise local geometry of the contact location; impact sounds at different surface locations are excited from different vibration patterns. In this benchmark task, we use or combine the object's visual, acoustic, and tactile observations for contact localization.

\vspace{-0.15in}
\paragraph{Task Definition.}
Given the object's mesh and different sensory observations of the contact position (visual images, impact sounds, or tactile readings), this task aims to predict the vertex coordinate of the surface location on the mesh where the contact happens. 

\begin{figure}
    \center
    \includegraphics[scale=0.26]{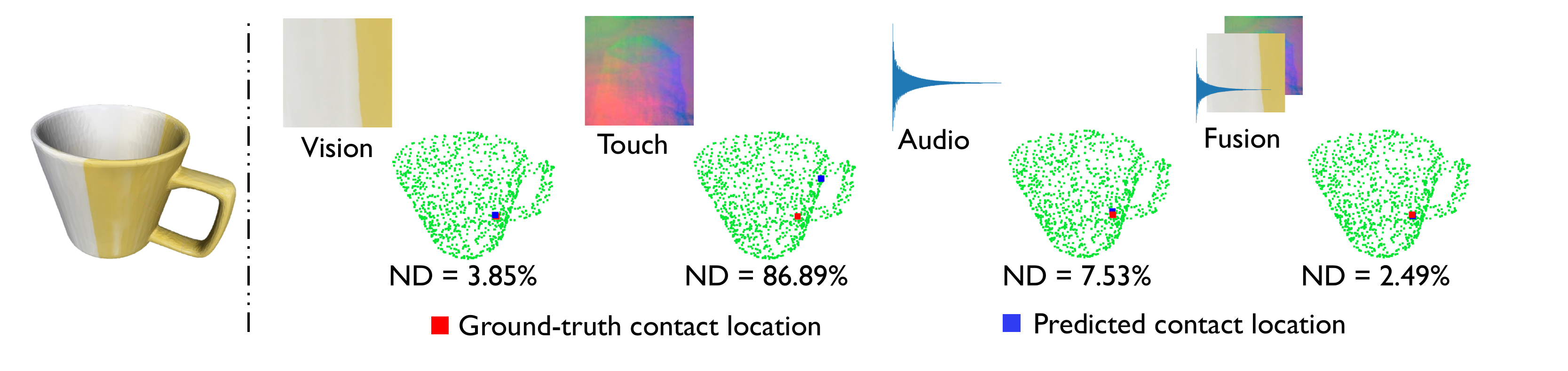}
    \vspace{-0.1in}
    \caption{Contact localization results for a ceramic mug object with our multisensory contact regression model. 
    }
    \label{fig:contact_localization_teaser_result}
\end{figure}

\vspace{-0.15in}
\paragraph{Evaluation Metrics and Baselines.}
We use the average Normalized Distance (ND) as our metric, which measures the distance between the predicted contact position and the ground-truth position normalized by the largest distance of two points on the object's surface. We evaluate an existing baseline Point Filtering~\cite{point_filtering,gao2022ObjectFolderV2}, where the contact position is recursively filtered out based on both the multisensory observations and the relative pose between consecutive contacts. This method performs very well but heavily relies on knowing the relative pose of the series of contacts, which might be a strong assumption in practice. Therefore, we also propose a new differentiable end-to-end learning baseline for contact localization---Multisensory Contact Regression (MCR), which takes the object mesh and multisensory observations as input to regress the contact position directly.

\vspace{-0.15in}
\paragraph{Teaser Results.}
Fig.~\ref{fig:contact_localization_teaser_result} shows an example result for a ceramic mug object with our MCR baseline. While vision and audio perform similarly, a single touch cannot easily locate where the contact is. Combining the three sensory modalities leads to the best result.

\subsection{Material Classification}\label{task:material}
\vspace{-0.05in}

\paragraph{Motivation.}
Material is an intrinsic property of an object, which can be perceived from different sensory modalities. For example, a ceramic object usually looks glossy, sounds crisp, and feels smooth. In this task, we predict an object's material category based on its multisensory observations. 

\vspace{-0.15in}
\paragraph{Task Definition.}
All objects are labeled by seven material types: ceramic, glass, wood, plastic, iron, polycarbonate, and steel. The task is formulated as a single-label classification problem. Given an RGB image, an impact sound, a tactile image, or their combination, the model must predict the correct material label for the target object.

\vspace{-0.15in}
\paragraph{Evaluation Metrics and Baselines.}
We report the classification accuracy and use two baselines: 1) ResNet~\cite{he2016deep} and 2) FENet~\cite{fenet}, which uses a different base architecture.
\vspace{-0.15in}
\paragraph{Teaser Results.}
We conduct material classification on both neural and real objects. Fusing different modalities largely improves the material classification accuracy. We also finetune the model trained on neural objects with only a few real-world measurements and achieve 6\% accuracy gain in classifying real objects.


\subsection{3D Shape Reconstruction}\label{task:3dreconstruction}
\begin{figure}
    \center
    \includegraphics[scale=0.4]{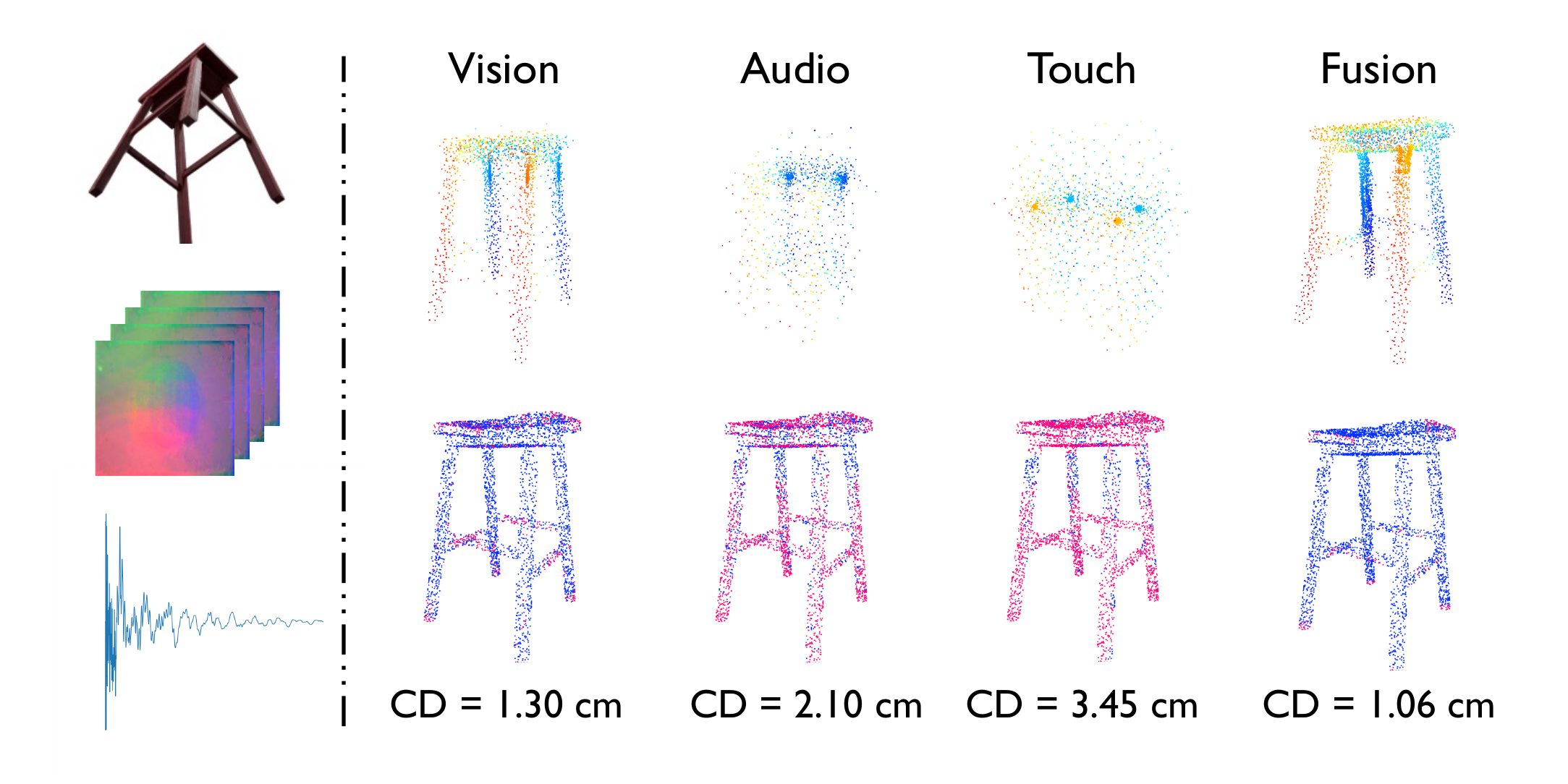}
    \caption{3D reconstruction results of a wooden chair object. The top/bottom row shows the point cloud reconstructions and the error over ground-truth points, respectively.
    Red indicates poorly-reconstructed areas; CD denotes Chamfer Distance.
    }
    \vspace{-0.1in}
    \label{fig:3d_recon_teaser_result}
\end{figure}
\paragraph{Motivation.}
While single-image shape reconstruction has been widely studied~\cite{zhang2017shape,choy20163d,mescheder2019occupancy,park2019deepsdf}, humans don't use vision alone to perceive the shape of objects. For example, we can touch an object's surface to sense its local details, or even knock and listen to the sound it makes to estimate its scale. The effective fusion of complementary multisensory information plays a vital role in 3D shape reconstruction, which we study in this benchmark task.

\vspace{-0.15in}
\paragraph{Task Definition.}
Given an RGB image of an object, a sequence of tactile readings from the object's surface, or a sequence of impact sounds of striking its surface locations, the task is to reconstruct the point cloud of the target object given combinations of these multisensory observations. This task is related to prior efforts on visuo-tactile 3D reconstruction~\cite{smith20203d,smith2021active,suresh2021efficient,rustler2022active}, but here we use all three sensory modalities and study their respective roles.

\vspace{-0.15in}
\paragraph{Evaluation Metrics and Baselines.}
We report Chamfer Distance~\cite{chamfer_distance} between the reconstructed and the ground-truth point cloud, a widely used metric to evaluate the quality of shape reconstruction. We use two state-of-the-art methods and a new transformer-based model as our baseline models: 1) Mesh Deformation Network (MDN)~\cite{smith20203d}, which is based on deforming the vertices of an initial mesh through a graph convolutional neural network, 2) Point Completion Network (PCN)~\cite{pcn,gao2022ObjectFolderV2}, which predicts the whole point cloud from latent features or incomplete point cloud constructed from local observations, and 3) Multisensory Reconstruction Transformer (MRT), which encodes multisensory data using a transformer-based architecture.

\vspace{-0.15in}
\paragraph{Teaser Results.}
For 3D reconstruction, our observation is that vision usually provides global yet coarse information, audio indicates the object's scale, and touch provides precise local geometry of the object's surface. Fig.~\ref{fig:3d_recon_teaser_result} shows an example of a wooden chair object. Both qualitative and quantitative results show that the three modalities make up for each other's deficiencies, and achieve the best reconstruction results when fused together.

\begin{figure}[t]
    \center
    \includegraphics[scale=0.39]{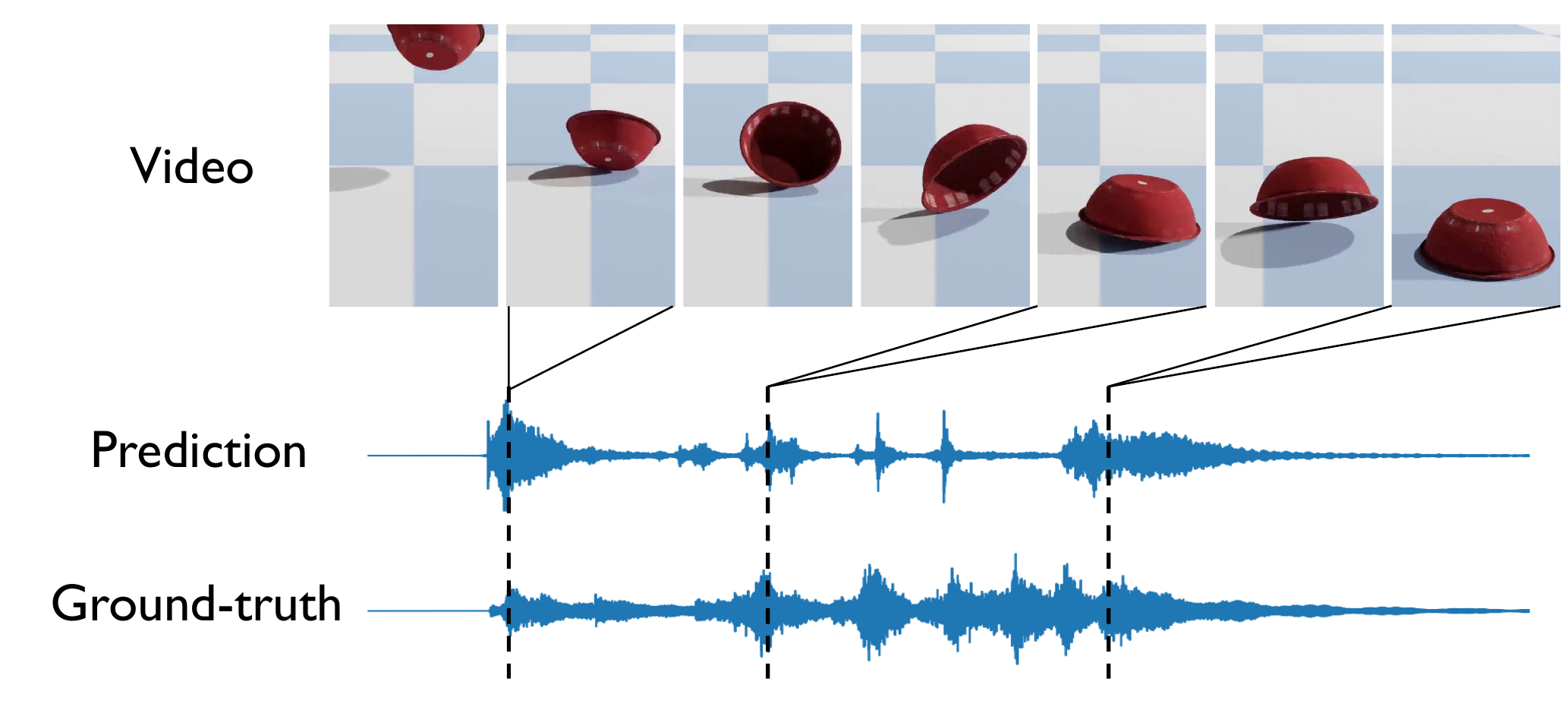}
    \vspace{-0.1in}
    \caption{Example results of sound generation for a falling steel bowl object with the RegNet~\cite{chen2020generating} baseline.    }
    \label{fig:sound_generation_teaser_result}
\end{figure}

\vspace{-0.05in}
\subsection{Sound Generation of Dynamic Objects}\label{task:sound_generation}

\paragraph{Motivation} Objects make unique sounds during interactions. When an object falls, we can anticipate how it sounds by inferring from its visual appearance and movement. In this task, we aim to generate the sound of dynamic objects based on videos displaying their moving trajectories. 

\vspace{-0.15in}
\paragraph{Task Definition.}
Given a video clip of a falling object, the goal of this task is to generate the corresponding sound based on the visual appearance and motion of the object. The generated sound must match the object's intrinsic properties (e.g., material type) and temporally align with the object's movement in the given video. This task is related to prior work on sound generation from in-the-wild videos~\cite{zhou2018visual,chen2020generating,iashin2021taming}, but here we focus more on predicting soundtracks that closely match the object dynamics.

\vspace{-0.15in}
\paragraph{Evaluation Metrics and Baselines.}
We use the following metrics for evaluating the sound generation quality: 1) STFT-Distance, which measures the Euclidean distance between the ground truth and predicted spectrograms, 2) Envelope Distance, which measures the Euclidean distance between the envelopes of the ground truth and the predicted signals, and 3) CDPAM~\cite{manocha2021cdpam}, which measures the perceptual audio similarity. We use two state-of-the-art methods as our baselines: RegNet~\cite{chen2020generating} and SpecVQGAN~\cite{iashin2021taming}. 

\vspace{-0.15in}
\paragraph{Teaser Results.}
Fig.~\ref{fig:sound_generation_teaser_result} shows an example of the predicted sound for a falling plate. We observe that the generated sound matches well with the ground-truth sound of the object perceptually, but it is challenging to predict the exact alignment that matches the object's motion.  

\begin{figure}
    \center
    \includegraphics[scale=0.385]{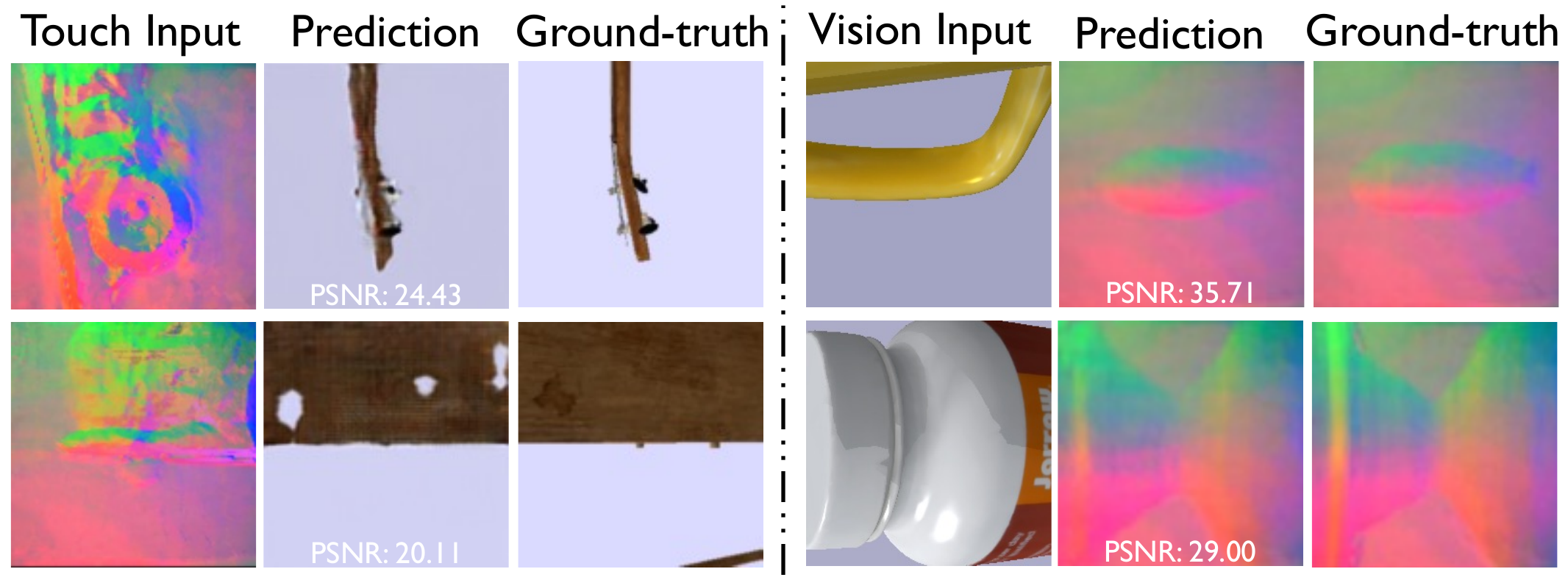}
    \vspace{-0.25in}
    \caption{Examples of Touch2Vision (left) and Vision2Touch (right) cross-generation results with the VisGel~\cite{li2019connecting} baseline.    }
    \label{fig:visual_tactile_generation_teaser_result}
\end{figure}
\vspace{-0.05in}
\subsection{Visuo-Tactile Cross-Generation}\label{task:visuo_tactile_generation}

\paragraph{Motivation.}
When we touch an object that is visually occluded (e.g., searching for a wallet from a backpack), we can often anticipate its visual textures and geometry merely based on the feeling on our fingertips. Similarly, we may imagine the feeling of touching an object purely from a glimpse of its visual appearance and vice-versa. To realize this intuition, we study the visuo-tactile cross-generation task initially proposed in~\cite{li2019connecting}. 

\vspace{-0.15in}
\paragraph{Task Definition.}
We can either predict touch from vision or vision from touch, leading to two subtasks: 1) Vision2Touch: Given an image of a local region on the object's surface, predict the corresponding tactile RGB image that aligns with the visual image patch in both position and orientation; and 2) Touch2Vision: Given a tactile reading on the object's surface, predict the corresponding local image patch where the contact happens. 

\vspace{-0.15in}
\paragraph{Evaluation Metrics and Baselines.}
Both the visual and tactile sensory data are represented by RGB images. Therefore, we evaluate the prediction performance for both subtasks using Peak Signal to Noise Ratio (PSNR) and Structural Similarity (SSIM) --- widely used metrics for assessing image prediction quality. We use two image-to-image translation methods as our baselines: 1) Pix2Pix~\cite{pix2pix}, which is a general-purpose conditional GAN framework, and 2) VisGel~\cite{li2019connecting}, which is a variant of Pix2Pix that is specifically designed for cross-sensory prediction. 

\vspace{-0.15in}
\paragraph{Teaser Results.}
Fig.~\ref{fig:visual_tactile_generation_teaser_result} shows some examples of visuo-tactile cross-generation.
 Very accurate touch signals can be reconstructed from local views of the objects, while visual image patches generated from tactile input tend to lose surface details. We suspect this is because different objects often share similar local patterns, making it ambiguous to invert visual appearance from a single tactile reading.

\begin{figure}[t]
    \centering
    \includegraphics[width=0.95\linewidth]{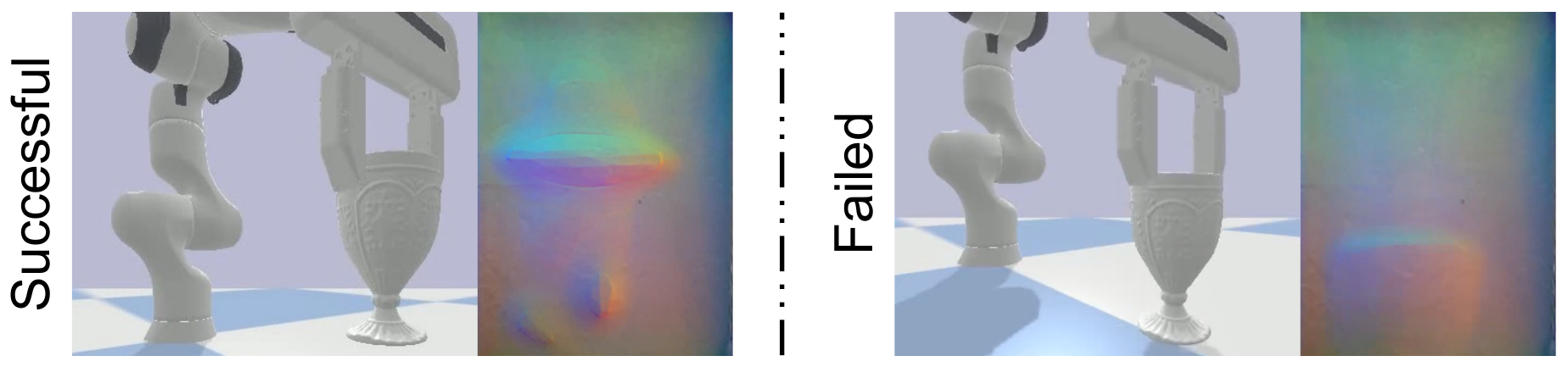}
    \vspace{0.15in}
    \begin{tabular}{cccc}
    \toprule
    \textsc{Chance} & V & T  & V + T \\
    \midrule
    50\% & 87.4\% & 93.4\% & 99.4\% \\
    \bottomrule
    \end{tabular}
    \vspace{-0.15in}
    \caption{Grasp stability prediction results with a wine glass. We show an example of a successful grasp (left) and one of a failed grasp (right). The table shows the prediction accuracy with V and T denoting using vision and/or touch, respectively.}
    \vspace{-0.1in}
    \label{fig:teaser_grasp_stability}
\end{figure}
\subsection{Grasp-Stability Prediction}\label{task:grasp_stability}

\paragraph{Motivation.}
Grasping an object is inherently a multisensory experience. When we grasp an object, vision helps us quickly localize the object, and touch provides an accurate perception of the local contact geometry. Both visual and tactile senses are useful for predicting the stability of robotic grasping, which has been studied in prior work with various task setups~\cite{calandra2017feeling,wang2020tacto,si2022grasp}.

\vspace{-0.15in}
\paragraph{Task Definition.} The goal is to predict whether a robotic gripper can successfully grasp and stably hold an object between its left and right fingers based on either an image of the grasping moment from an externally mounted camera, a tactile RGB image obtained from the GelSight robot finger, or their combination. The grasp is considered failed if the grasped object slips by more than 3 cm.

\vspace{-0.15in}
\paragraph{Evaluation Metrics and Baselines.} We report the accuracy of grasp stability prediction. We implement TACTO~\cite{wang2020tacto} as the baseline method, which uses a ResNet-18~\cite{he2016deep} network for feature extraction from the visual and tactile RGB images to predict the grasp stability.

\vspace{-0.15in}
\paragraph{Teaser Results.}
We show a successful and a failed grasp for a wine glass in Fig.~\ref{fig:teaser_grasp_stability}. Vision and touch are both helpful in predicting grasp stability, and combining the two sensory modalities leads to the best result. 

\begin{figure}[t]
    \centering
    \includegraphics[width=0.95\linewidth]{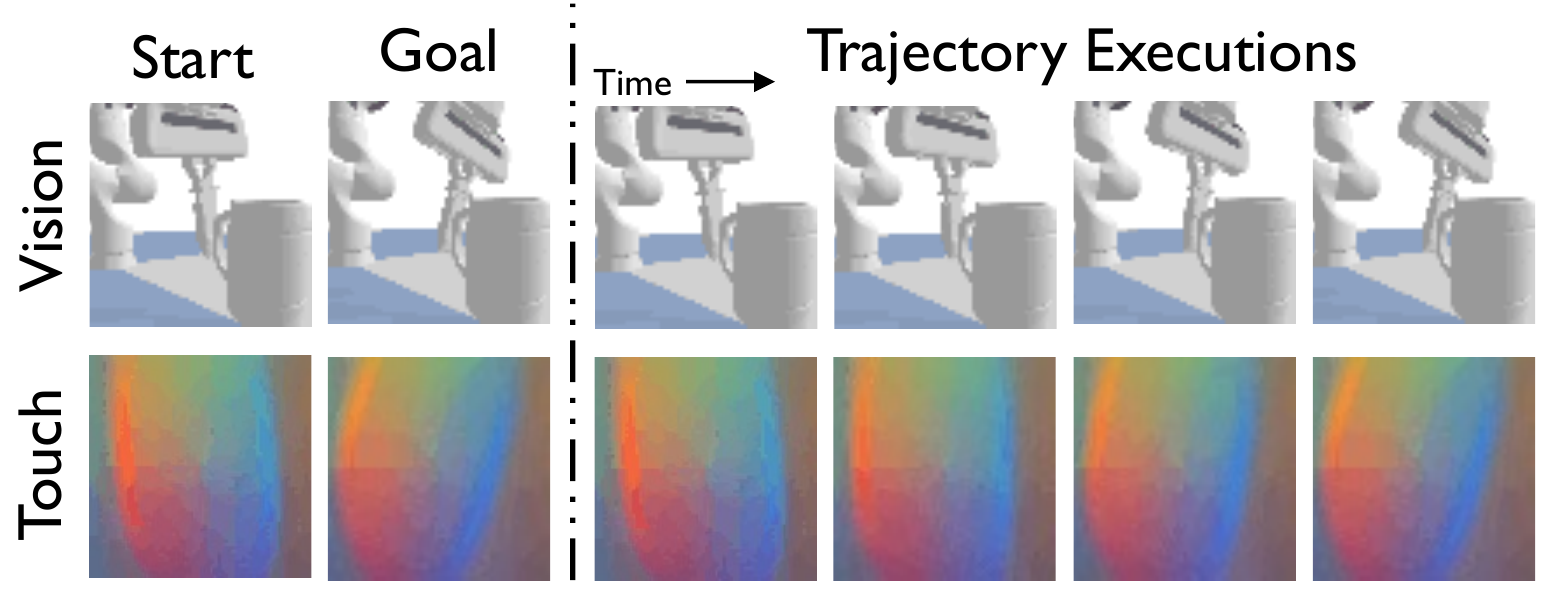}
    \vspace{0.15in}
    \begin{tabular}{lcccc}
    \toprule
    & V & T  & V + T \\
    \midrule
    SR~$\uparrow$ & 0.86 & 0.83 & 0.88\\
    AE~$\downarrow$ ($^{\circ}$)  & 0.38 & 0.56 & 0.34\\
    \bottomrule
    \end{tabular}
    \vspace{-0.25in}
    \caption{Contact refinement results of a wooden cup object. From left to right, we show the start and goal observations for both vision (top) and touch (bottom), and the actual trajectory executions. The table shows the success rate (SR) and the angle error (AE) for using vision (V), touch (T), or its combination.
    }
    \label{fig:teaser_contact_refinement}
\end{figure}

\subsection{Contact Refinement}\label{task:refinement}

\paragraph{Motivation.} 
When seeing a cup, we can instantly analyze its shape and structure, and decide to put our fingers around its handle to lift it. We often slightly adjust the orientations of our fingers to achieve the most stable pose for grasping. For robots, locally refining how it contacts an object is of great practical importance. We define this new task as \emph{contact refinement}, which can potentially be a building block for many dexterous manipulation tasks.

\vspace{-0.15in}
\paragraph{Task Definition.} Given an initial pose of the robot finger, the task is to change the finger's orientation to contact the point with a different target orientation. Each episode is defined by the following: the contact point, the start orientation of the robot finger along the vertex normal direction of the contact point, and observations from the target finger orientation in the form of either a third view camera image, a tactile RGB image, or both. We use a continuous action space over the finger rotation dimension. The task is successful if the finger reaches the target orientation within 15 action steps with a tolerance of 1$^{\circ}$.

\vspace{-0.15in}
\paragraph{Evaluation Metrics and Baselines.}
We evaluate using the following metrics: 1) success rate (SR),  which is the fraction of successful trials, and 2) average Angle Error (AE) across all test trials. Model Predictive Control (MPC)~\cite{finn2017deep,ebert2018visual,tian2019manipulation} has been shown to be a powerful framework for planning robot actions. Therefore, we implement Multisensory-MPC as our baseline, which uses SVG~\cite{villegas2019high} for future frame prediction, and Model Predictive Path Integral Control (MPPI)~\cite{grady2016mppi} for training the control policy.

\vspace{-0.15in}
\paragraph{Teaser Results.}
Fig.~\ref{fig:teaser_contact_refinement} shows a trajectory execution example for using both vision and touch. We can obtain an 88\% success rate and average angle error of 0.17$^{\circ}$ by combining both modalities using our Multisensory-MPC baseline. 

\begin{figure}[t!]
    \centering
    \includegraphics[width=0.95\linewidth]{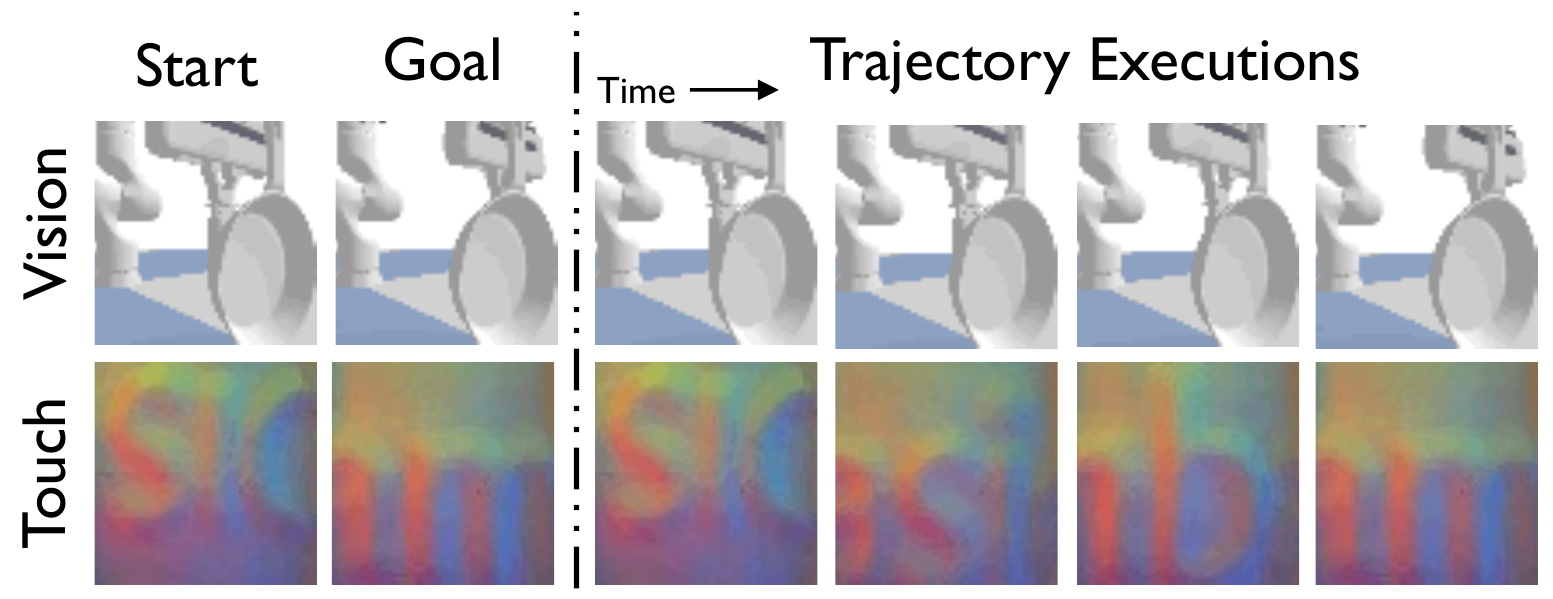}
    \vspace{0.15in}
    \begin{tabular}{lccc}
    \toprule
    & V & T  & V + T \\
    \midrule
    SR~$\uparrow$ & 0.26 & 0.54 & 0.80\\
    PE~$\downarrow$ ($mm$)  & 1.77 & 3.32 & 0.84\\
    \bottomrule
    \end{tabular}
    \vspace{-0.15in}
    \caption{Trajectory executions examples for surface traversal with an iron pan. The table shows the success rate (SR) and average position error (PE) for using vision (V) and/or touch (T).    }
    \vspace{-0.1in}
    \label{fig:teaser_surface_traveral}
\end{figure}
\subsection{Surface Traversal}\label{task:surface_traversal}

\paragraph{Motivation.}
When a robot's finger first contacts a position on an object, it may not be the desired surface location. Therefore, efficiently traversing from the first contact point to the target location is a prerequisite for performing follow-up actions or tasks. We name this new task \emph{surface traversal}, where we combine visual and tactile sensing to efficiently traverse to the specified target location given a visual and/or tactile observation of the starting location.

\vspace{-0.2in}
\paragraph{Task Definition.}
Given an initial contacting point, the goal of this task is to plan a sequence of actions to move the robot finger horizontally or vertically in the contact plane to reach another target location on the object's surface. Each episode is defined by the following: the initial contact point, and observations of the target point in the form of either a third-view camera image, a tactile RGB image, or both. The task is successful if the robot finger reaches the target point within 15 action steps with a tolerance of 1 mm.

\vspace{-0.2in}
\paragraph{Evaluation Metrics and Baselines.} We report the following two metrics: 1) success rate (SR), and 2) average position error (PE), which is the average distance between the final location of the robot finger on the object's surface and the target location. We use the same Multisensory-MPC baseline as in the contact refinement task.

\vspace{-0.2in}
\paragraph{Teaser Results.}
Fig.~\ref{fig:teaser_surface_traveral} shows the surface traversal results with an iron pan, where the back of the pan has a sequence of letters. The Multisensory-MPC model can successfully traverse from the start location to the goal location. We observe significant gains when combining vision and touch, achieving a success rate of 80\%.

\begin{figure}[t!]
    \centering
    \includegraphics[width=0.93\linewidth]{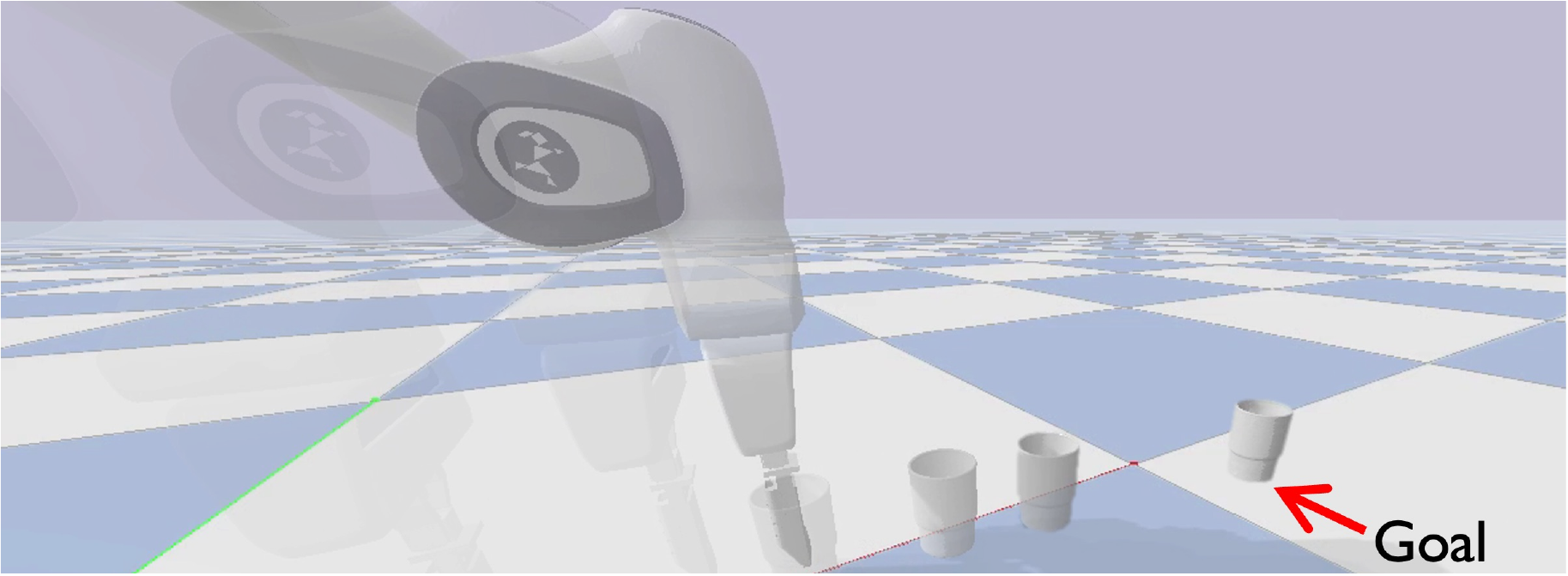}
    \vspace{0.15in}
    \begin{tabular}{lcccc}
    \toprule
    & V & T  & V + T \\
    \midrule
    PE~$\downarrow$ ($cm$)  & 23.81 & 21.76 & 17.63\\
    \bottomrule
    \end{tabular}
    \vspace{-0.15in}
    \caption{Examples of dynamic pushing. The table shows the average position error (PE) for using vision (V) and/or touch (T) with a rinsing cup.}
    \vspace{-0.1in}
\label{fig:dynamic_pushing_teaser_result}
\end{figure}
\subsection{Dynamic Pushing}\label{task:dynamic_pushing}

\paragraph{Motivation.} To push an object to a target location, we use vision to gauge the distance and tactile feedback to control the force and orientation. For example, in curling, the player sees and decides on the stone's target, holds its handle to push, and lightly turns the stone in one direction or the other upon release. Both visual and tactile signals play a crucial role in a successful delivery. We name this task \emph{dynamic pushing}, which is related to prior work on dynamic adaptation for pushing~\cite{evans2022context} with only vision.

\vspace{-0.15in}
\paragraph{Task Definition.} Given example trajectories of pushing different objects together with their corresponding visual and tactile observations, the goal of this task is to learn a forward dynamics model that can quickly adapt to novel objects with a few contextual examples. With the learned dynamics model, the robot is then tasked to push the objects to new goal locations.

\vspace{-0.15in}
\paragraph{Evaluation Metrics and Baselines.}
We report the average position error (PE) across all test trials. For the baseline, we use a ResNet-18 network for feature extraction and a self-attention mechanism for modality fusion to learn the forward dynamics model. We use a sampling-based optimization algorithm (i.e., cross-entropy method~\cite{de2005tutorial}) to obtain the control signal.

\vspace{-0.15in}
\paragraph{Teaser Results.}
Fig.~\ref{fig:dynamic_pushing_teaser_result} shows an example of pushing a novel test object to a new goal location. Vision and touch are both useful for learning object dynamics, and combining the two sensory modalities leads to the best results.
\vspace{-0.05in}
\section{Conclusion}
We presented the \name, a suite of 10 benchmark tasks centered around object recognition, reconstruction, and manipulation to advance research on multisensory object-centric learning. We also introduced \namereal, the first dataset that contains all visual, acoustic, and tactile real-world measurements of 100 real household objects. We hope our new dataset and benchmark suite can serve as a solid building block to enable further research and innovations in multisensory object modeling and understanding.


\vspace{-0.12in}
\begin{small}
\paragraph{Acknowledgments.} We thank Samuel Clarke, Miaoya Zhong, Mark Rau, Hong-Xing Yu, and Samir Agarwala for helping with the data collection, and thank Doug James, Zilin Si, Fengyu Yang, and Yen-Yu Chang for helpful discussions. This work is in part supported by Amazon, NSF CCRI \#2120095, NSF RI \#2211258, ONR MURI N00014-22-1-2740, AFOSR YIP FA9550-23-1-0127, the Stanford Institute for Human-Centered AI (HAI), the Toyota Research Institute (TRI), Adobe, and Meta.
\end{small}

\clearpage
\appendix

\addcontentsline{toc}{section}{Appendix} %
\part{Appendix} %
\parttoc %

\addcontentsline{toc}{section}{Appendices}
\renewcommand{\thesubsection}{\Alph{subsection}}

\subsection{Summary of Benchmark Tasks}\label{sec:grasp_stability_supp}

 We introduce a suite of 10 benchmark tasks for multisensory object-centric learning, centered around \emph{object recognition}, \emph{object reconstruction}, and \emph{object manipulation}. Table~\ref{tab:benchmark_summary} illustrates which of these tasks can be performed with simulated data, real-world data, or both. All 10 tasks can be done in simulation. We have obtained results using \namereal for four tasks, including cross-sensory retrieval, contact localization, material classification, and 3D shape reconstruction. For sound generation of dynamic objects and visuo-tactile cross-generation, sim2real transfer is not feasible due to the large sim-real gap, and the collected data in \namereal is not directly applicable to these two tasks. Performing real-world versions of these two tasks may require collecting real datasets tailored for these two tasks. For manipulation, each task needs nontrivial effort for real-world robot deployment. For example, prior work~\cite{si2022grasp} has made a dedicated effort to make sim2real transfer possible for grasp stability prediction with careful calibration of their physics simulator of robot dynamics, contact model, and the tactile optical simulator with real-world data. We provide some tentative guidelines on sim2real transfer in Sec.~\ref{sec:sim2real_supp} and hope our open-sourced simulation framework can encourage future exploration of sim2real transfer for these four tasks.

\begin{table*}[h]
\centering
\begin{tabular}{ccccccccccc}
     & \multicolumn{3}{c}{Object Recognition}                    & \multicolumn{3}{c}{Object Reconstruction}                 & \multicolumn{4}{c}{Object Manipulation}                                             \\ 
     \cmidrule(lr){2-4} \cmidrule(lr){5-7}\cmidrule(lr){8-11}
     & \multicolumn{1}{c}{CSR} & \multicolumn{1}{c}{CL} & MC & \multicolumn{1}{c}{3DSR} & \multicolumn{1}{c}{SGoDO} & VTCG & \multicolumn{1}{c}{GSP} & \multicolumn{1}{c}{CR} & \multicolumn{1}{c}{ST} & DP \\ 
     \midrule
sim  & \multicolumn{1}{c}{\small\ding{51}}  & \multicolumn{1}{c}{\small\ding{51}}  &  \multicolumn{1}{c}{\small\ding{51}} & \multicolumn{1}{c}{\small\ding{51}}  & \multicolumn{1}{c}{\small\ding{51}}  & \multicolumn{1}{c}{\small\ding{51}}  & \multicolumn{1}{c}{\small\ding{51}}  & \multicolumn{1}{c}{\small\ding{51}}  & \multicolumn{1}{c}{\small\ding{51}}  & \multicolumn{1}{c}{\small\ding{51}}  \\
real & \multicolumn{1}{c}{\small\ding{51}}  & \multicolumn{1}{c}{\small\ding{51}}  &  \multicolumn{1}{c}{\small\ding{51}} & \multicolumn{1}{c}{\small\ding{51}}  & \multicolumn{1}{c}{\small\ding{56}}  & \multicolumn{1}{c}{\small\ding{56}}  & \multicolumn{1}{c}{?}  & \multicolumn{1}{c}{?}  & \multicolumn{1}{c}{?}  &   ? \\
\end{tabular}
\caption{We summarize all 10 benchmark tasks to show whether they can be performed with simulated data, real-world data, or both. CSR, CL, MC, 3DSR, SGoDO, VTCG, GSP, CR, ST, DP denote cross-sensory retrieval, contact localization, material classification, 3D shape reconstruction, sound generation of dynamic objects, visuo-tactile cross-generation, grasp-stability prediction, contact refinement, surface traversal, and dynamic pushing, respectively.}
\label{tab:benchmark_summary}
\end{table*}

\subsection{Cross-Sensory Retrieval}\label{sec:retrieval_supp}

In this section, we detail the cross-sensory retrieval benchmark task definition and settings, baseline methods and evaluation metrics, and the experiment results.

\subsubsection{Task Definition and Settings}

Cross-sensory retrieval requires the model to take one sensory modality as input and retrieve the corresponding data of another modality. For instance, given the sound of striking a mug, the ``audio2vision'' model needs to retrieve the corresponding image of the mug from a pool of images of hundreds of objects. In this benchmark, each sensory modality (vision, audio, touch) can be used as either input or output, leading to $9$ sub-tasks.

Specifically, we sample $100$ instances from each modality of each object, resulting in two instance sets $S_A$ and $S_B$. Next, we pair the instances from both modalities, which is done by the Cartesian Product:
\begin{equation}
    P(i)=S_A(i) \times S_B(i),
\end{equation} where $i$ is the object index and $P$ is the set of instance pairs. For each object, given modality A and modality B (A and B can be either vision, touch or audio), the goal of cross-sensory retrieval is to minimize the distance between the representations of sensory observations from the same object while maximizing those from different objects. In our experiments, we randomly split the objects from \namecvpr into train/val/test splits of 800/100/100 objects, and split the 10 instances of each object from \namereal into 8/1/1.

\begin{table*}[t!]
\centering
    \begin{tabular}{c c c c c c c}
        \toprule
        Input & Retrieved & RANDOM & CCA~\cite{cca} & PLSCA~\cite{plsca}  & DSCMR~\cite{dar} & DAR~\cite{dscmr} \\
        \midrule
        \multirow{3}*{Vision} & Vision (different views) & $1.00$ & $55.52$ & $82.43$ & $82.74$ & $\textbf{89.28}$ \\
        ~ & Audio & $1.00$ & $19.56$ & $11.53$ & $9.13$ & $\textbf{20.64}$  \\
        ~ & Touch & $1.00$ & $6.97$ & $6.33$ & $3.57$ & $\textbf{7.03}$  \\
        \midrule
        \multirow{3}*{Audio} & Vision & $1.00$ & $\textbf{20.58}$ & $13.37$ & $10.84$ & $20.17$ \\
        ~ & Audio (different vertices) & $1.00$ & $70.53$ & $\textbf{80.77}$ & $75.45$ & $77.80$ \\
        ~ & Touch & $1.00$ & $5.27$ & $\textbf{6.96}$ & $5.30$ & $6.91$ \\
        \midrule
        \multirow{3}*{Touch} & Vision & $1.00$ & $8.50$ & $6.25$ & $4.92$ & $\textbf{8.80}$ \\
        ~ & Audio & $1.00$ & $6.18$ & $7.11$ & $6.15$ & $\textbf{7.77}$ \\
        ~ & Touch (different vertices) & $1.00$ & $28.06$ & $52.30$ & $51.08$ & $\textbf{54.80}$ \\
        \bottomrule
    \end{tabular}
\caption{Experiment results of the cross-sensory retrieval task using neural objects from \namecvpr. We evaluate the performance using mean Average Precision (mAP).}
\label{tab:retrieval_results}
\end{table*}


\subsubsection{Baselines and Evaluation Metrics}
We use the following four state-of-the-art methods as our baselines:
\begin{itemize}
    \item Canonical Correlation Analysis (CCA)~\cite{cca}: CCA is a traditional method to analyze the correlation between two datasets, which reduces the data dimension by a linear projection. Specifically, in the testing process, we leverage a ResNet~\cite{he2016deep} pre-trained with instance recognition on the $800$ objects in the training set to extract the features from the multisensory data. Next, the features are projected into a unified representation space by CCA.
    \item Partial Least Squares (PLSCA)~\cite{plsca}: we follow the same feature extracting process as CCA, except that the final projection step is replaced with PLSCA, which combines CCA with the partial least squares (PLS).
    \item Deep Supervised Cross-Modal Retrieval (DSCMR)~\cite{dscmr}: DSCMR is proposed to conduct image-text retrieval by minimizing three losses: 1) the discrimination loss in the label space, which utilizes the ground-truth category label as supervision, 2) the discrimination loss in the shared space, which measures the similarity of the representations of different modalities, helping the network to learn discriminative features, and 3) the inter-modal invariance loss, which eliminates the cross-modal discrepancy by minimizing the distance between the representations of instances from the same category. We follow similar settings in our experiments. 
    \item Deep Aligned Representations (DAR)~\cite{dar}: the DAR model is trained with both a model transfer loss and a ranking pair loss. The model transfer loss utilizes a teacher model to train the DAR student model, which in our setting is a ResNet~\cite{he2016deep} model pretrained on our data. The student model is trained to predict the same class probabilities as the teacher model, which is measured by the KL-divergence. The ranking pair loss is used to push the instances from the same object closer in the shared space, and push those from different objects apart from each other.
\end{itemize}

In the retrieval process, we set each instance in the input sensory modality as the query, and the instances from another sensory are retrieved by ranking them according to cosine similarity. Next, the Average Precision (AP) is computed by considering the retrieved instances from the same object as positive and others as negative. Finally, the model performance is measured by the mean Average Precision (mAP) score, which is a widely-used metric for evaluating retrieval performance.

\subsubsection{Experiment Results}
Table~\ref{tab:retrieval_results} and Table~\ref{tab:retrieval_results_real} show the cross-sensory retrieval results on the neural objects from \namecvpr and the real objects from \namereal, respectively.
We have the following observation from the experiment results. Compared with the modality that encodes local information (touch), the modalities encoding global information (vision and audio) are more informative to perform cross-sensory retrieval. This is because different objects may share similar local tactile features, but their visual images and impact sounds are discriminative (e.g., a steel cup and a ceramic bowl). 

\begin{table*}
\centering
    \begin{tabular}{c c c c c c c}
        \toprule
        Input & Retrieved & RANDOM & CCA~\cite{cca} & PLSCA~\cite{plsca}  & DSCMR~\cite{dar} & DAR~\cite{dscmr} \\
        \midrule
        \multirow{3}*{Vision} & Vision (different views) & $3.72$ & $30.60$ & $60.95$ & $\textbf{81.27}$ & $81.00$ \\
        ~ & Audio & $3.72$ & $12.05$ & $27.12$ & $\textbf{68.34}$ & $66.92$  \\
        ~ & Touch & $3.72$ & $6.29$ & $9.77$ & $\textbf{64.91}$ & $39.46$  \\
        \midrule
        \multirow{3}*{Audio} & Vision & $3.72$ & $12.41$ & $30.54$ & $\textbf{67.16}$ & $64.35$ \\
        ~ & Audio (different vertices) & $3.72$ & $27.40$ & $55.75$ & $\textbf{72.59}$ & $68.79$ \\
        ~ & Touch & $3.72$ & $5.38$ & $11.66$ & $\textbf{54.55}$ & $33.00$ \\
        \midrule
        \multirow{3}*{Touch} & Vision & $3.72$ & $6.40$ & $11.46$ & $\textbf{64.86}$ & $41.18$ \\
        ~ & Audio & $3.72$ & $5.57$ & $13.89$ & $\textbf{55.37}$ & $37.30$ \\
        ~ & Touch (different vertices) & $3.72$ & $21.16$ & $27.97$ & $\textbf{66.09}$ & $41.42$ \\
        \bottomrule
    \end{tabular}
\caption{Experiment results of the cross-sensory retrieval task using real objects from \namereal. We evaluate the performance using mean Average Precision (mAP).}
\label{tab:retrieval_results_real}
\end{table*}


\subsection{Contact Localization}\label{sec:contactlocalization_supp}

In this section, we detail the contact localization benchmark task definition and settings, baseline methods and evaluation metrics, and the experiment results.

\subsubsection{Task Definition and Settings}

Given the object's mesh and different sensory observations of the contact position (visual images, impact sounds, or tactile readings), the multisensory contact localization task aims to predict the vertex coordinate of the surface location on the mesh where the contact happens. More formally, the task can be defined as follows: given a visual patch image $V$ (i.e., a visual image near the object's surface) and/or a tactile reading $T$ and/or an impact sound $S$, and the shape of the object $P$ (represented by a point cloud), the model needs to localize the contact position $C$ on the point cloud. The task objective can be described as:
\begin{equation}
    \underset{\theta}{\min}\left\{ \textrm{Dist}\left( f_{\theta}\left( V,T,S,P \right) ,C \right) \right\},
\end{equation}
where $f_{\theta}$ denotes the model for contact localization.

Specifically, we manually choose $50$ objects with rich surface features from the dataset, and sample $1,000$ contacts from each object. The sampling strategy is based on the surface curvature. We assume that the curvature of each vertex is subject to a uniform distribution. The average value of vertex curvatures is computed at first, and the vertices with curvatures that are far from the average value are sampled with higher probability (i.e., the vertices with more special surface patterns are more likely to be sampled).

In the experiments, we randomly split the $1,000$ instances of each object into train/val/test splits of 800/190/10, respectively. Similarly, in the real experiments, we choose 53 objects from \namereal and randomly split the instances of each object by 8:1:1.

\begin{table*}[]
    \centering
    \begin{tabular}{c c c c c c c c}
    \toprule
        Method & Vision & Touch & Audio & Vision+Touch & Vision+Audio & Touch+Audio & Vision+Touch+Audio \\
    \midrule
        RANDOM & $47.32$& $47.32$& $47.32$& $47.32$& $47.32$& $47.32$& $47.32$\\
        
        Point Filtering~\cite{point_filtering} & $-$ & $4.21$ & $\textbf{1.45}$ & $-$ & $-$ & $3.73$ & $-$\\
        
        MCR & $5.03$ & $23.59$ & $4.85$ & $4.84$ & $\textbf{1.76}$ & $3.89$ & $1.84$\\
    \bottomrule
    \end{tabular}
    \caption{Results of Multisensory Contact Localization on \namecvpr 2.0. We use average Normalized Distance (ND) as the evaluation metric. The numbers are all in percent ($\%$).}
    \label{tab:contact_localization_result}
\end{table*}

\begin{table}[]
    \centering
    \begin{tabular}{c c c c c}
    \toprule
        Method & Vision & Touch & Audio & Fusion \\
    \midrule
        RANDOM & $50.57$& $50.57$& $50.57$& $50.57$\\
        
        MCR & $12.30$ & $32.03$ & $35.62$ & $\textbf{12.00}$\\
    \bottomrule
    \end{tabular}
    \caption{Results of Multisensory Contact Localization on \namereal. We use average Normalized Distance (ND) as the evaluation metric. The numbers are all in percent ($\%$). The Point Filtering method requires obtaining touch/audio data at arbitrary points on the object's surface, which is not available for the collected real object data in \namereal. Thus this method is not included in this table.}
\label{tab:contact_localization_result_real}
\end{table}

\subsubsection{Baselines and Evaluation Metrics}

We evaluate an existing method as our first baseline and also develop a new end-to-end differentiable baseline model for contact localization:
\begin{itemize}
    \item Point Filtering~\cite{point_filtering,gao2022ObjectFolderV2}: this represents a typical pipeline for contact localization, where the contact positions are recurrently filtered out based on both the multisensory input data and the relative displacements between the contacts of a trajectory. Each trajectory contains $8$ contacts, at each iteration of the filtering process, possible contact positions are generated on the object surface, and the positions whose touch or audio features are similar to the input data are kept with higher probability. As a result, the predictions gradually converge into a small area, which is treated as the final prediction. We only evaluate on the final contact of each trajectory. This method predicts very accurate results but heavily relies on the relative displacements between the contacts instead of the multisensory information. Furthermore, the filtering process is not differentiable, thus not being able to be optimized end-to-end.
    \item Multisensory Contact Regression (MCR): in order to solve the limitations of the point filtering method, we propose this novel differentiable baseline for contact localization. In this method, the model takes the object point cloud and multisensory data as input and directly regresses the contact position.
\end{itemize}

The models' performance is evaluated by the average Normalized Distance (ND), which is the distance between the predicted contact position and the ground-truth position normalized by the largest distance between the two vertices on the object mesh. The reason for adopting this metric is to fairly evaluate objects with different scales.

\subsubsection{Experiment Results}

We have the following two key observations from the results shown in Table~\ref{tab:contact_localization_result}. Firstly, compared with touch, contact localization using vision and audio achieves much better results, because they provide more global information and suffer from less ambiguity (i.e., different positions on the object may share similar surface tactile features, resulting in large ambiguity). Secondly, though MCR performs worse than point filtering, it shows promising results that are close to the point filtering results even with a simple network architecture. This shows the great potential of end-to-end contact localization methods. In Table~\ref{tab:contact_localization_result_real}, we show the contact localization results on \namereal.
\subsection{Material Classification}\label{sec:material_supp}

In this section, we detail the multisensory material classification benchmark task definition and settings, baseline methods and evaluation metrics, and the experiment results.

\subsubsection{Task Definition and Settings}

All objects are labeled by seven material types: ceramic, glass, wood, plastic, iron, polycarbonate, and steel. The multisensory material classification task is formulated as a single-label classification problem. Given an RGB image, an impact sound, a tactile image, or their combination, the model must predict the correct material label for the target object. The $1,000$ objects are randomly split into train: validation: test = $800: 100: 100$, and the model needs to generalize to new objects during the testing process. 
Furthermore, we also conduct a cross-object experiment on \namereal to test the Sim2Real transferring ability of the models, in which the $100$ real objects are randomly split into train: validation: test = $60: 20: 20$.

\begin{table}[t]
    \centering
    \begin{tabular}{c c c c c}
    \toprule
        Method & Vision & Touch & Audio & Fusion \\
    \midrule
        ResNet~\cite{he2016deep} & $91.89$ & $74.36$ & $94.91$ & $96.28$ \\
        FENet~\cite{fenet} & $\textbf{92.25}$ & $\textbf{75.89}$ & $\textbf{95.80}$ & $\textbf{96.60}$\\
    \bottomrule
    \end{tabular}
    \caption{Results on Multisensory Material Classification. We evaluate the model performance by top-1 accuracy. The numbers are all in percent ($\%$).}
    \label{tab:material_classification_result}
\end{table}

\begin{table}[t]
    \centering
    \begin{tabular}{l c}
    \toprule
        Method & Accuracy$\uparrow$ \\
    \midrule
        ResNet~\cite{he2016deep} w/o pretrain & $45.25$ \\
        ResNet~\cite{he2016deep} & $\textbf{51.02}$ \\
    \bottomrule
    \end{tabular}
    \caption{Transfer learning results of material classification on \namereal. We evaluate the model performance by top-1 accuracy. The numbers are all in percent ($\%$).}
    \label{tab:material_classification_result_real}
\end{table}

\subsubsection{Baselines and Evaluation Metrics}

We use the following two methods as our baselines:
\begin{itemize}
    \item ResNet~\cite{he2016deep}: we finetune the ResNet backbone pretrained on ImageNet~\cite{deng2009imagenet}, which is considered as a naive baseline.
    \item Fractal Encoding Network~\cite{fenet}: the Fractal Encoding (FE) module is originally proposed for texture classification task, which is a trainable module that encode the multi-fractal texture features. We apply this module to the ResNet baseline, enabling it to encode the multisensory object features. 
\end{itemize}
We evaluate the model performance by top-1 accuracy, which is a standard metric for classification tasks.

\subsubsection{Experiment Results}
Tab.~\ref{tab:material_classification_result} shows the comparison between the two baselines on the simulation data. The Fractal Encoding module brings about $1\%$ improvement. Touch modality performs much worse than vision and audio due to its lack of global information, and the fusion of the three modalities leads to the best performance.

Tab.~\ref{tab:material_classification_result_real} shows the Sim2Real experiment results. We evaluate the performance of ResNet~\cite{he2016deep} with/without the pre-training on neural objects. Results show that pre-training on the simulation data brings about $6\%$ improvement.

\begin{table*}[t!]
    \centering
    \begin{tabular}{c c c c c c c c}
    \toprule
        Method & Vision & Touch & Audio & Vision+Touch & Vision+Audio & Touch+Audio & Vision+Touch+Audio \\
    \midrule
        MDN \cite{smith20203d} & $4.02$ & $3.88$ & $5.04$ & $3.19$ & $4.05$ & $3.49$ & $\textbf{2.91}$\\
        PCN \cite{pcn} & $2.36$ & $3.81$ & $3.85$ & $2.30$ & $2.48$ & $3.27$ & $\textbf{2.25}$\\
        MRT & $2.80$ & $4.12$& $5.01$& $\textbf{2.78}$& $3.13$& $4.28$& $3.08$\\
    \bottomrule
    \end{tabular}
    \caption{Results of Multisensory 3D Reconstruction on \namecvpr 2.0, we use chamfer distance (cm) as the metric to measure the model performance. Lower is better.}
    \label{tab:3d_recon_result}
\end{table*}

\begin{table*}[t!]
    \centering
    \begin{tabular}{c c c c c c c c}
    \toprule
        Method & Vision & Touch & Audio & Vision+Touch & Vision+Audio & Touch+Audio & Vision+Touch+Audio \\
    \midrule
        MRT &$1.17$&$1.04$&$1.64$&$0.96$&$1.50$&$1.12$&$\textbf{0.95}$\\
    \bottomrule
    \end{tabular}
    \caption{Results of Multisensory 3D Reconstruction on \namereal. We use Chamfer Distance (cm) as the metric to measure the model performance. Lower is better.}
    \label{tab:3d_recon_result_real}
\end{table*}

\subsection{3D Shape Reconstruction}\label{sec:3dreconstruction_supp}

In this section, we detail the multisensory 3D reconstruction benchmark task definition and settings, baseline methods and evaluation metrics, and the experiment results.

\subsubsection{Task Definition and Setting}

Given an RGB image of an object $V$, a sequence of tactile readings $T$ from the object's surface, or a sequence of impact sounds $S$ of striking $N$ surface locations of the object, the task is to reconstruct the 3D shape of the whole object represented by a point cloud given combinations of these multisensory observations. The procedure can be denoted as:
\begin{equation}
    \underset{\theta}{\min}\left\{ \textrm{Dist}\left( f_{\theta}\left( V,T,S \right) ,\mathrm{Points}_{\mathrm{GT}} \right) \right\},
\end{equation}
where $f_{\theta}$ represents the model for multisensory 3D reconstruction and $\mathrm{Points}_{\mathrm{GT}}$ represents the ground-truth point cloud. 
This task is related to prior efforts on visuo-tactile 3D reconstruction~\cite{smith20203d,smith2021active,suresh2021efficient,rustler2022active}, but here we include all three sensory modalities and study their respective roles.

For the visual RGB images, tactile RGB images, and impact sounds used in this task, we respectively sample $100$ instances around each object (vision) or on its surface (touch and audio). In all, given the $1,000$ objects, we can obtain $1,000 \times 100=100,000$ instances for vision, touch, and audio modality, respectively. In the experiments, we randomly split the $1,000$ objects as train: validation: test = $800:100:100$, meaning that the models need to generalize to new objects during testing.
Furthermore, we also test the model performance on \namereal by similarly splitting the $100$ objects as train: validation: test = $60: 20: 20$.

\subsubsection{Baseline and Evaluation Metrics}

We first use two state-of-the-art methods as our baselines, and we further develop a transformer-based baseline model for 3D Reconstruction:
\begin{itemize}
    \item Mesh Deformation Network (MDN)~\cite{smith20203d}: this method first predicts local charts from the tactile images and combine them with the initial global chart. Next, the model deforms the combined chart based on vision and/or audio signal by an iterative process, in which the touch consistency is ensured (i.e., the local charts remain unchanged). The final prediction is a deformed chart, which is then transformed into a point cloud by sampling on its surface.
    
    \item Point Completion Network (PCN)~\cite{pcn}: this method infers the complete point cloud based on the coarse global point cloud (predicted by vision and audio) and/or detailed local point cloud (predicted by touch).

    \item Multisensory Reconstruction Transformer (MRT): when touch is used in the reconstruction process, the previous two methods require first predicting local point clouds/meshes based on tactile readings. In our setting, the prediction is done by transforming the depth maps of the tactile readings into local point clouds. 
    However, accurate depth maps can only be obtained in the simulation setting. In our setting for real capture, only the tactile RGB images are captured, thus making it impossible for MDN and PCN to perform 3D reconstruction using tactile data of \namereal.
    To solve this limitation, we propose a new model, Multisensory Reconstruction Transformer (MRT), as a new baseline model. In this method, the model directly takes a sequence of tactile RGB images as input and encodes them into a latent vector by a transformer encoder.
    Specifically, the images are first forwarded into a ResNet~\cite{he2016deep} model to obtain a sequence of features. Next, each feature is concatenated with a learnable token that attends to all features in the attention layer. Finally, the concatenated sequence is sent into the transformer encoder and the output feature (i.e., the first token of the output sequence) is decoded into the point cloud prediction by a simple MLP. 
    The method can also encode a sequence of impact sounds in a similar way.
    
\end{itemize}

The performance of each baseline is measured by Chamfer Distance (CD), which calculates the distance between two point clouds by:
\begin{equation}
    \mathrm{CD}=\frac{1}{S_1}\sum_{x\in S_1}{\underset{y\in S_2}{\min}\left\| x-y \right\| _{2}^{2}}+\frac{1}{S_2}\sum_{y\in S_2}{\underset{x\in S_1}{\min}\left\| y-x \right\| _{2}^{2}},
\end{equation}
where $S_1$ and $S_2$ are two point clouds.

\subsubsection{Experiment Results}
Table~\ref{tab:3d_recon_result} and Table~\ref{tab:3d_recon_result_real} show the experiment results on both simulation and real settings. We can obtain some key findings from the results. Firstly, if only one single modality is used, vision does much better than the other two modalities. This shows that the global information captured by the visual signal is most important for 3D reconstruction.

Secondly, when different modalities are combined, the tactile readings can significantly improve the reconstruction from either vision or audio, while the audio data can only benefit the reconstruction from touch in most cases. Moreover, when all three modalities are combined, the best results are achieved in most experiments. We suspect this results from the following different characteristics of the three modalities: 1) vision data provides global information (shape and scale) of the objects, while only a few local surface details can be obtained from a single image; 2) tactile readings contain very detailed local information of the touched areas but miss the global context; 3) audio data only provides rough scale information (i.e., the size of objects), while it is hard to infer fine-grained details of the objects from audio.

\subsection{Sound Generation of Dynamic Objects}\label{sec:video_sound_generation_supp}

In this section, we detail the sound generation of dynamic objects task definition and settings, baseline methods and evaluation metrics, and the experiment results.

\subsubsection{Task Definition and Settings}

Given a video clip of a falling object, the goal of this task is to generate the corresponding sound based on the visual appearance and motion of the object. The generated sound must match the object's intrinsic properties (e.g., material type) and temporally align with the object's movement in the given video. This task is related to prior work on sound generation from in-the-wild videos~\cite{zhou2018visual,chen2020generating,iashin2021taming}, but here we focus more on predicting soundtracks that closely match the object dynamics.


We adopt a process similar to~\cite{jin2022neuralsounda} to generate the data for this task. Firstly, the physical simulation is performed in the Pybullet~\cite{coumans2016pybullet} simulator. We put the object above the floor in the simulator and randomly set an initial velocity. The object is then released and will have contact with the floor, during which the object pose, contact positions, and contact forces are recorded. Secondly, we query the \emph{ObjectFile} implicit representation network of the object with the contact positions and forces to obtain the impact sounds. The sounds are then temporally aligned into a single waveform, which is the ground-truth audio. Finally, we render the video using the Blender software, which generates the video according to the object pose at each frame.

Specifically, we choose $500$ objects with reasonable scales, and $10$ videos are generated for each object. We split the $10$ videos into train/val/test splits of 8/1/1.

\subsubsection{Baselines and Evaluation Metrics}
We use two state-of-the-art methods as the baselines:
\begin{itemize}
    \item RegNet~\cite{chen2020generating}: in this work, a novel module called audio forwarding regularizer is proposed to solve the incorrect mapping between the video frames and sound. During training, both the video frames and ground-truth sound is used to predict the spectrogram. The regularizer only takes the ground-truth sound as the input and encode it into a latent feature, which is considered as ``visual-irrelevant'' information. The model then predicts the spectrogram according to both the ``visual-relevant'' information provided by the video frames and the ``visual-irrelevant'' information. This architecture helps the model correctly map the visual signal to the audio signal. During testing, the regularizer is turned off, meaning the model should predict the spectrogram based on merely the video frames. With the proper regularizer size, the model can capture useful and correct information from the visual signal.
    
    \item SpecVQGAN~\cite{iashin2021taming}: in this work, a more complex framework is proposed to generate the visually relevant sounds. A transformer is trained to autoregressively generate codebook representations based on frame-wise video features. The representation sequence is then decoded into a spectrogram. 
\end{itemize}
For the waveform prediction, we pretrain a MelGAN~\cite{kumar2019melgan} vocoder on our dataset, which is used to reconstruct the temporal information of the spectrogram, transforming it into a sound waveform.

To comprehensively measure the sound generation quality, we evaluate the model performance by three metrics that respectively computes the distance between the prediction and ground-truth in spectrogram space, waveform space, and latent space: 1) STFT-Distance, 2) Envelope Distance, and 3) CDPAM~\cite{manocha2021cdpam}.

\subsubsection{Experiment Results}

The results in Table~\ref{tab:sound_generation_result} show that RegNet model performs slightly better than the SpecVQGAN model under all of the three metrics, though the SpecVQGAN model is larger and more complex. This is probably because the transformer model used in  SpecVQGAN requires more data to be adequately trained. See the Supp. video for the qualitative comparison results.
\begin{table}
    \centering
    \begin{tabular}{c c c c}
    \toprule
        Method & STFT$\downarrow$ & Envelope$\downarrow$ & CDPAM$\downarrow$\\
    \midrule
        RegNet~\cite{chen2020generating} & $\textbf{0.010}$ & $\textbf{0.036}$ & $\textbf{5.65}\times 10^{-5}$ \\
        SpecVQGAN~\cite{iashin2021taming} & $0.034$ & $0.042$ & $5.92\times 10^{-5}$ \\
    \bottomrule
    \end{tabular}
    \caption{Results of generating object sound from video.}
    \label{tab:sound_generation_result}
\end{table}
\subsection{Visuo-Tactile Cross-Generation}\label{sec:visuo_tactile_generation_supp}

In this section, we detail the visuo-tactile cross-generation task definition and settings, baseline methods and evaluation metrics, and the experiment results.

\subsubsection{Task Definition and Settings}


The visuo-tactile cross-generation task is originally proposed in~\cite{li2019connecting}. The task requires the model to reconstruct the tactile image from the visual input or vice versa. 
Similarly, we define the following two subtasks: 1) Vision2Touch: Given an image of a local region on the object's surface, predict the corresponding tactile RGB image that aligns with the visual image patch in both position and orientation; and 2) Touch2Vision: Given a tactile reading on the object's surface, predict the corresponding local image patch where the contact happens. 

Specifically, we choose $50$ objects with rich tactile features and reasonable size, and sample $1,000$ visuo-tactile image pairs on each of them. This results in $50\times 1,000=50,000$ image pairs. We conduct both cross-contact and cross-object experiments by respectively splitting the $1,000$ visuo-tactile pairs of each object into train: validation: test = $800: 100: 100$ and splitting the $50$ objects into train: validation: test = $40: 5: 5$. The two settings require the model to generalize to new areas or new objects during testing.

\subsubsection{Baselines and Evaluation Metrics}

We use the following two state-of-the-art methods as the baselines:
\begin{itemize}
    \item Pix2Pix~\cite{pix2pix}: Pix2Pix is a general-purpose framework for image-to-image translation. The model is optimized by both the L1 loss and a GAN loss, which respectively make the generated image similar to the target and looks realistic. In our benchmark, we utilize Pix2Pix to predict the images in both directions.
    \item VisGel~\cite{li2019connecting}: VisGel is a modification of Pix2Pix, which is designed for visuo-tactile cross generation specifically. This work indicates that the huge domain gap between vision and touch makes it extremely difficult to conduct generation in both directions. To solve this problem, VisGel adds a reference image to the input, which in their setting is a global image of the initial scene or the empty GelSight reading. Similarly, we also add reference images to the input of both directions in our setting. The visual reference is an image of the whole object, showing the global shape and texture of the object, and the tactile reference is the background of our GelSight sensor.
\end{itemize}
The prediction of the task is a generated image, thus should be evaluated by metrics that assess the image quality. We adopt two metrics in our benchmark: Peak Signal to Noise Ratio (PSNR) and Structural Similarity (SSIM), which are widely used for evaluating image generation tasks.

\subsubsection{Experiment Results}
The experiment results are shown in Tab.~\ref{tab:visuo_tactile_generation_result_cross_contact} and Tab.~\ref{tab:visuo_tactile_generation_result_cross_object}. We can have the following two key observations from the results. Firstly, generating tactile images from visual images is much easier than the reversed direction. We observe that very accurate tactile signals can often be reconstructed, while many of the generated visual images look hardly reasonable, even if the reference images are provided. This is probably due to the fact that different objects may share similar tactile patterns, making it difficult to infer a visual signal from a single tactile reading. Secondly, the reference information used in VisGel brings huge improvement. Providing the global visual signal or empty tactile reading helps the model bridge the domain gap between vision and touch, making it able to produce much more realistic images. The improvement is also clearly shown by the quantitative results measured by both metrics.
\begin{table}
    \centering
    \begin{tabular}{c c c c c}
    \toprule
        \multirow{2}*{Method} & \multicolumn{2}{c}{Vision $\rightarrow$ \text{Touch}} & \multicolumn{2}{c}{Touch $\rightarrow$ \text{Vision}} \\
        & PSNR$\uparrow$ & SSIM$\uparrow$ & PSNR$\uparrow$ & SSIM$\uparrow$ \\
    \midrule
        pix2pix~\cite{pix2pix} & $22.85$ & $0.71$ & $9.16$ & $0.28$ \\
        VisGel~\cite{li2019connecting} & $\textbf{29.60}$ & $\textbf{0.87}$ & $\textbf{14.56}$ & $\textbf{0.61}$ \\
    \bottomrule
    \end{tabular}
    \caption{Cross-contact experiment results of visuo-tactile generation on $50$ selected objects.}
    \label{tab:visuo_tactile_generation_result_cross_contact}
\end{table}

\begin{table}
    \centering
    \begin{tabular}{c c c c c}
    \toprule
        \multirow{2}*{Method} & \multicolumn{2}{c}{Vision $\rightarrow$ \text{Touch}} & \multicolumn{2}{c}{Touch $\rightarrow$ \text{Vision}} \\
        & PSNR$\uparrow$ & SSIM$\uparrow$ & PSNR$\uparrow$ & SSIM$\uparrow$ \\
    \midrule
        pix2pix~\cite{pix2pix} & $18.91$ & $0.63$ & $7.03$ & $0.12$ \\
        VisGel~\cite{li2019connecting} & $\textbf{25.91}$ & $\textbf{0.82}$ & $\textbf{12.61}$ & $\textbf{0.38}$ \\
    \bottomrule
    \end{tabular}
    \caption{Cross-object experiment results of visuo-tactile generation on $50$ selected objects.}
    \label{tab:visuo_tactile_generation_result_cross_object}
\end{table}
\subsection{Grasp Stability Prediction}\label{sec:grasp_stability_supp}

\begin{table*}[t]
\centering
\begin{tabular}{ccccc}
\toprule
Chance & ImageNet~\cite{deng2009imagenet} & \namecvpr 2.0~\cite{gao2022ObjectFolderV2} & Touch and Go~\cite{yang2022touch} & \namereal \\ 
\midrule
  56.1\%     &  73.0\%  &  69.4\%        &    78.1\%  &        \textbf{84.9\%} \\
\bottomrule
\end{tabular}
\caption{Transfer learning results. We show the grasp stablity prediction results on the dataset from~\cite{calandra2017feeling} by pre-training on ImageNet~\cite{deng2009imagenet}, and other tactile datasets, including \namecvpr 2.0~\cite{gao2022ObjectFolderV2}, Touch and Go~\cite{yang2022touch}, and our new \namereal  dataset.}
\label{tab:transfer_learning}
\end{table*}

In this section, we detail the grasp stability prediction benchmark task definition and settings, baseline methods and evaluation metrics, and the experiment results.

\subsubsection{Task Definition and Setting}

Both visual and tactile senses are useful for predicting the stability of robotic grasping, which has been studied in prior work with various task setups~\cite{calandra2017feeling,wang2020tacto,si2022grasp}. The goal of this task is to predict whether a robotic gripper can successfully grasp and stably hold an object between its left and right fingers based on either an image of the grasping moment from an externally mounted camera, a tactile RGB image obtained from the GelSight robot finger, or their combination. 

More specifically, we follow the settings of \cite{wang2020tacto,si2022grasp} on setting up the grasping pipeline. The robot takes the specified grasping configuration, including the target orientation and height, moves to the specified location, and closes the gripper with a certain speed and force. After the gripper closes entirely, we record the tactile images from the GelSight sensor as the tactile observations and record the images from the third-view camera as the visual observations. These observations are used as input to our grasp stability prediction model. Then, the robot attempts to lift the object 18 cm to the ground. The grasp is considered failed if the grasped object slips by more than 3 cm. Otherwise, it's considered successful.

We generate 10,000 grasping examples for each object. We balance the success and failure cases to be around 1:1. We randomly split the dataset into 9,000 samples for training and 1,000 samples for testing. We choose 5 different objects with different materials and shapes suitable for the grasping task.

\subsubsection{Baseline and Evaluation Metrics}

We use TACTO~\cite{wang2020tacto} as the baseline method, which uses a ResNet-18~\cite{he2016deep} network for feature extraction from the visual and tactile RGB images to predict the grasp stability. We use cross-entropy loss to train the binary classification network with different sensory inputs. We report the grasp stability prediction accuracy as the evaluation metric.

\subsubsection{Experiment results}

Table~\ref{Table:grasp_stability_supp_results} shows the results on 5 representative objects from \namereal. The results consistently suggest that vision and touch both play a crucial role in predicting grasp stability. Combining the two sensory modalities leads to the best performance. 

In addition, we use this task as a case study to evaluate representations pre-trained on \namereal compared to \namecvpr 2.0~\cite{gao2022ObjectFolderV2} and the Touch and Go dataset~\cite{yang2022touch}, which are the largest simulated dataset and human-collected visuo-tactile dataset in the literature, respectively. We also compare with a baseline that performs supervised pre-training on ImageNet~\cite{deng2009imagenet}. 

Following the settings in~\cite{yang2022touch}, we learn tactile representations with visuo-tactile contrastive multiview coding~\cite{tian2020contrastive}, and then use the setup and dataset of~\cite{calandra2017feeling} for evaluating grasp stability prediction. We extract visuo-tactile pairs from the videos we record with the third-view camera and the tactile sensor during data collection. We extract 3 pairs in the last 0.5 seconds for each point, leading to 10.6K visuo-tactile pairs in total. 

Table~\ref{tab:transfer_learning} shows the results. We quote the baseline results directly from~\cite{yang2022touch}. Pre-training on \namereal outperforms prior datasets by a large margin, demonstrating the value and potential of transfer learning using our dataset.

\begin{table*}
\small
\begin{tabular}{lccccc}
\toprule
     & {\includegraphics[height=.1\textwidth]{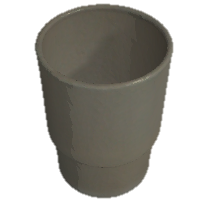}}  & {\includegraphics[height=.1\textwidth]{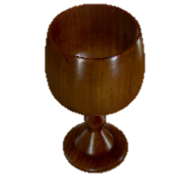}}  & {\includegraphics[height=.1\textwidth]{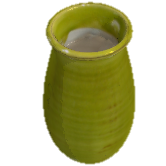}}  & {\includegraphics[height=.1\textwidth]{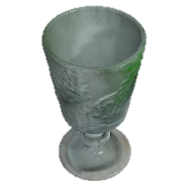}}  &
     {\includegraphics[height=.1\textwidth]{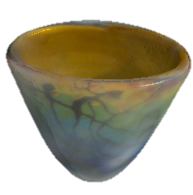}} 
     \\ 
\midrule
     Vision & 87.4\%  & 77.3\% & 81.7\% & 79.2\% & 77.7\% \\ 
     Touch & 90.1\% & 81.0\% & 89.0\% & 84.3\% & 89.1\% \\
     Vision + Touch & 92.0\% & 88.9\% & 93.8\% & 85.5\% & 90.6\% \\
\bottomrule
\end{tabular}
\caption{Results on grasp stability prediction. We report the prediction accuracy with vision and/or touch.}
\label{Table:grasp_stability_supp_results}
\end{table*}

\subsection{Contact Refinement}\label{sec:contact_refinement_supp}

In this section, we detail the contact refinement benchmark task definition and settings, baseline methods and evaluation metrics, and the experiment results.

\begin{table*}[ht]
\small
\begin{tabular}{lcccccccccccc}
\toprule
\multirow{2}{*}{Modalities} & \multicolumn{2}{c}{\includegraphics[height=.1\textwidth]{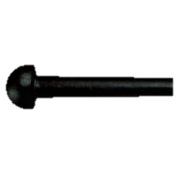}} & \multicolumn{2}{c}{ \includegraphics[height=.1\textwidth]{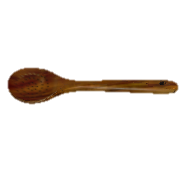}} &
\multicolumn{2}{c}{\includegraphics[height=.1\textwidth]{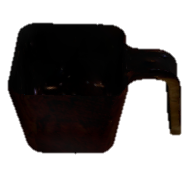}} & \multicolumn{2}{c}{\includegraphics[height=.1\textwidth]{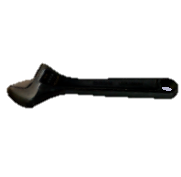}} &
\multicolumn{2}{c}{\includegraphics[height=.1\textwidth]{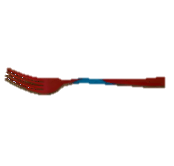}} 
 \\ 
\cmidrule(lr){2-3}\cmidrule(lr){4-5} \cmidrule(lr){6-7} \cmidrule(lr){8-9} \cmidrule(lr){10-11}
                  & SR~$\uparrow$   & AE~$\downarrow$     & SR~$\uparrow$    & AE~$\downarrow$  & SR~$\uparrow$  & AE~$\downarrow$   & SR~$\uparrow$    & AE~$\downarrow$     & SR~$\uparrow$    & AE~$\downarrow$   \\ \midrule
Vision   &    0.91     &   0.31         &  0.96   &  0.24  &  0.88   &  0.36  & 0.95 & 0.24  & 0.91 & 0.33         \\ 
Touch        &  0.86      &   0.41     &     0.93      &  0.34   &     0.88       &  0.37   &  0.95  & 0.28 &  0.91  & 0.32   \\
Vision + Touch        &  0.94      &   0.26     &     0.97       &  0.21   &     0.92       &  0.27   &  0.96 & 0.21 &  0.93  & 0.24   \\
\bottomrule
\end{tabular}
\caption{Results on contact refinement. We report the success rate (SR) and average angle error (AE) for using vision and/or touch for 5 objects from our \namereal dataset. $\uparrow$ denotes higher is better, $\downarrow$ denotes lower is better. }
\label{Table:contact_refinement_supp_results}
\end{table*}

\subsubsection{Task Definition and Setting}

Given an initial pose of the robot finger, the goal of the contact refinement task is to change the finger's orientation to contact the point with a different target orientation. Each episode is defined by the following: the contact point, the start orientation of the robot finger along the vertex normal direction of the contact point, and observations from the target finger orientation in the form of either a third view camera image, a tactile RGB image, or both. We use a continuous action space over the finger rotation dimension. The task is successful if the finger reaches the target orientation within 15 action steps with a tolerance of 1$^{\circ}$. Based on the object category, we choose a local region of interest (RoI) for the robot to touch (e.g., the handle of the cup). The discrete Gaussian curvature~\cite{cohen2003restricted} of the RoI should be larger than 0. The robot will randomly select a point in that local region and touches that point with a random finger orientation. Then, the robot samples actions from a Gaussian distribution, and repeats the sampled action four times before it samples the next action. We set the area of RoI to be around 5 $cm^2$ and sampled 600 points for training and 100 points for testing. 

\subsubsection{Baseline and Evaluation Metrics}

Model Predictive Control (MPC)~\cite{finn2017deep,ebert2018visual,tian2019manipulation} has been shown to be a powerful framework for planning robot actions. Therefore, we implement Multisensory-MPC as our baseline, which uses SVG~\cite{villegas2019high} for future frame prediction, and Model Predictive Path Integral Control (MPPI)~\cite{grady2016mppi} for training the control policy.  

To train the video prediction model, We collect 600 trajectories for training and 100 trajectories for evaluation. Each trajectory has 20 steps. We train a separate model for vision and touch for each object. During training, we randomly sample a sequence of 14 steps, from which we condition on the first 2 frames and predict 12 future frames. For MPC, we use MPPI with a squared error objective, which calculates the pixel-wise error and samples actions based on it. The horizon length is 10 steps, which means the model will sample an action sequence of length 10 into the future. The robot should finish the task within 15 steps, beyond which we consider the task fails.

\subsubsection{Experiment results}

In the main paper, we have shown a trajectory execution example for using both vision and touch. Table~\ref{Table:contact_refinement_supp_results} shows the contact refinement results of 5 objects from the \namereal dataset. We can see that vision and touch are both very useful for contact refinement. Combining the two modalities leads to the best success rate and can refine more accurately to the target location.

\subsection{Surface Traversal}\label{sec:surface_traversal_supp}

In this section, we detail the surface traversal benchmark task definition and settings, baseline methods and evaluation metrics, and the experiment results.

\begin{table*}[ht]
\small
\begin{tabular}{lcccccccccccc}
\toprule
\multirow{2}{*}{Modalities} & \multicolumn{2}{c}{\includegraphics[height=.1\textwidth]{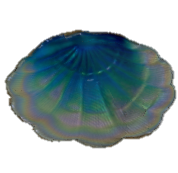}} & \multicolumn{2}{c}{ \includegraphics[height=.1\textwidth]{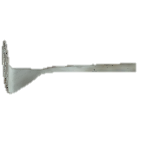}} &
\multicolumn{2}{c}{\includegraphics[height=.1\textwidth]{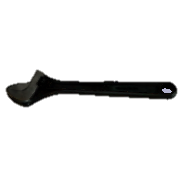}} & \multicolumn{2}{c}{\includegraphics[height=.1\textwidth]{Figures/robotics_tasks_objects/53.png}} &
\multicolumn{2}{c}{\includegraphics[height=.1\textwidth]{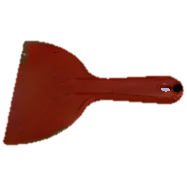}} 
 \\ 
\cmidrule(lr){2-3}\cmidrule(lr){4-5} \cmidrule(lr){6-7} \cmidrule(lr){8-9} \cmidrule(lr){10-11}
                  & SR~$\uparrow$    & PE~$\downarrow$    & SR~$\uparrow$   & PE~$\downarrow$ & SR~$\uparrow$ & PE~$\downarrow$  & SR~$\uparrow$   & PE~$\downarrow$    & SR~$\uparrow$   & PE~$\downarrow$  \\ \midrule
Vision   &    0.03     &     6.47       &  0.24   &  2.40  & 0.27   & 2.23  & 0.28 & 2.06  & 0.18 & 2.78         \\ 
Touch        &  0.20      &   6.88     &     0.08       &  6.51   &     0.05     &  9.91   &  0.06 & 8.16 &  0.06 & 7.93   \\
Vision + Touch        &  0.18      &   5.95     &     0.36       &  1.75   &     0.20       &  6.88   &  0.18  & 2.36 & 0.23  & 3.42   \\
\bottomrule
\end{tabular}
\caption{Results on surface traversal. We report the success rate (SR) and average position error (PE) in $mm$ for using vision and/or touch for 5 objects from our \namereal dataset. $\uparrow$ denotes higher is better, $\downarrow$ denotes lower is better.}
\label{Table:surface_traversal_supp_results}
\end{table*}

\subsubsection{Task Definition and Setting}

Given an initial contacting point, the goal of this task is to plan a sequence of actions to move the robot finger horizontally or vertically in the contact plane to reach another target location on the object's surface. Each episode is defined by the following: the initial contact point, and observations of the target point in the form of either a third-view camera image, a tactile RGB image, or both. The task is successful if the robot finger reaches the target point within 15 action steps with a tolerance of 1 mm. We follow a similar data generation protocol as the contact refinement task. Based on the object's category and geometry, we select a local region of interest (RoI) for the robot to traverse. The discrete Gaussian curvature~\cite{cohen2003restricted} of the RoI should be larger than 0 and less than 0.01. The robot starts at a random location in that region and samples actions from a Gaussian distribution along two directions. The robot repeats the sampled action four times before it samples the next action. The number of sampled trajectories is proportional to the area of RoI with 50 trajectories per $1 \ cm^2$ for training and 5 trajectories per $1 \ cm^2$ for testing.

\subsubsection{Baseline and Evaluation Metrics}

Similar to the contact refinement task, we implement Multisensory-MPC as our baseline, which uses SVG~\cite{villegas2019high} for future frame prediction, and Model Predictive Path Integral Control (MPPI)~\cite{grady2016mppi} for training the control policy. We evaluate using the following metrics: 1) success rate (SR),  which is the fraction of successful trials, and 2) average Angle Error (AE) across all test trials. For the video prediction model, we collect 2,000 trajectories for training and 200 trajectories for evaluation. Then, we follow the same control pipeline as in the contact refinement task.

\subsubsection{Experiment results}

Table~\ref{Table:surface_traversal_supp_results} shows the results of surface traversal with 5 objects from the \namereal dataset. Generally, we observe that the performance of this task is very object-dependent. Vision provides global information about the object, while touch offers precise contact geometry. Therefore, combining the two modalities often leads to more accurate traversal results. However, our current  Mulisensory-MPC model cannot make the most of the benefit from the two modalities, sometimes leading to worse results compared to the performance of a single modality.

\subsection{Dynamic Pushing}\label{sec:dynamic_pushing_supp}

In this section, we detail the dynamic pushing benchmark task definition and settings, baseline methods and evaluation metrics, and the experiment results.

\begin{table*}
\small
\begin{tabular}{lccccc}
\toprule
     & {\includegraphics[height=.1\textwidth]{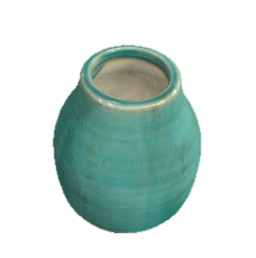}}  & {\includegraphics[height=.1\textwidth]{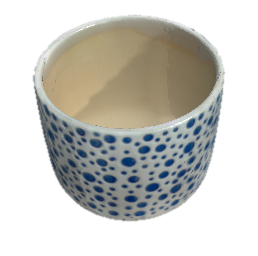}}  & {\includegraphics[height=.1\textwidth]{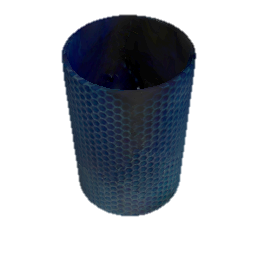}}  & {\includegraphics[height=.1\textwidth]{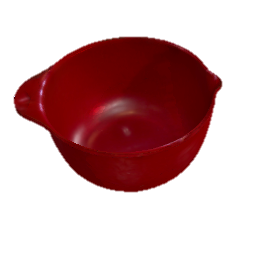}}  &
     {\includegraphics[height=.1\textwidth]{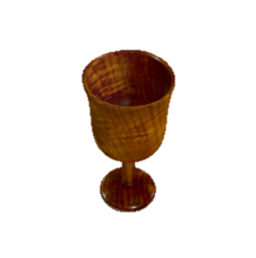}} 
     \\ 
\midrule
     Vision & 24.04  & 18.37 & 18.26 & 19.02 & 23.39  \\ 
     Touch &  30.06 & 30.79 & 34.65 & 26.22 & 26.12\\
     Vision + Touch & 30.30 & 22.00 & 18.88 & 18.94 & 19.25 \\
\bottomrule
\end{tabular}
\caption{Results on dynamic pushing. We report the average position error (PE) in $cm$  for using vision and/or touch for 5 objects from the \namereal dataset.}
\label{Table:dynamic_pushing_supp_results}
\vspace{-0.1in}
\end{table*}

\subsubsection{Task Definition and Settings}

Given example trajectories of pushing different objects together with their corresponding visual and tactile observations, the goal of this task is to learn a forward dynamics model that can quickly adapt to novel objects with a few contextual examples. With the learned dynamics model, the robot is then tasked to push the objects to new goal locations. 

More specifically, the object is initialized at a fixed location in front of the robot. We specify the angle between the line passing through the center-of-mass of the object and the of the gel on the GelSight sensor and the x-axis. This angel defines the pushing direction. We also specify a pushing distance, which is the distance along the pushing direction. The pushing speed stays the same for all trials. With these two parameters, the robot can push the object to some positions in front of it.

We select 16 cylinder-shaped objects for training and collect 200 trials for each object. We vary the object's mass and friction coefficients every 10 trials. For evaluation, we select 6 unseen objects with different geometry, mass, and friction coefficients and run 500 trials for each object.

\subsubsection{Baseline and Evaluation Metrics}

For our baseline model, we use a ResNet-18 network for feature extraction and a three-layer MLP to learn the forward dynamics model. We use a sampling-based optimization algorithm (i.e., cross-entropy method (CEM)~\cite{de2005tutorial}) to obtain the control signal. During training, we encode a feature vector by taking in observations from 3 trials of the object with the same mass and friction and use that feature vector to train the dynamics model. The dynamics model takes in the feature vector, the angle, and the pushing distance to predict the final position of that object. During testing, we use CEM as the control policy with L2 distance between the predicted location and the goal location as the cost. Then, by specifying the goal location, the dynamics model can predict the corresponding action that reaches the goal, represented by the pushing angle and the pushing distance. We use the average position error (PE) across all test trials as our metric.

\subsubsection{Experiment results}

Table~\ref{Table:dynamic_pushing_supp_results} shows the results. We can see that vision and touch are both useful for learning object dynamics. Combining the two sensory modalities leads to the best results for objects with simple surface geometry.

\subsection{Sim2Real Guidelines}\label{sec:sim2real_supp}

In this section, we provide some tentative guidelines on potentially transferring from simulation to real-world regarding the four robotic manipulation tasks as a reference for future work, including optical calibration and elastic deformation calibration.

\subsubsection{Optical Calibration}

The GelSight tactile images are rendered with a state-of-the-art simulation framework, Taxim\cite{si2021taxim}. Taxim uses a lookup table to map the contact shapes to tactile images. Following the pipeline in \cite{si2021taxim}, we have made similar attempts to press a ball with a radius of 4mm over the elastomer surface and manually locate contact areas in the tactile images. The polynomial lookup table can be calibrated with the collected data.

\subsubsection{Elastic Deformation Calibration}
To eliminate the gap between sim-to-real transfer, we also need to calibrate the physics parameter of contact dynamics using real-world data. The elastic deformation can be simplified into two parts: normal and lateral displacements. Taxim uses linear mapping to characterize the relationship between the indentation displacement and the normal force. Using a force gauge stand, we can collect a set of force-displacement pairs to fit the physics parameter along the normal direction. For lateral displacements, we haven't found a standard and general procedure to calibrate the simulator with real-world data for all four tasks. A potential approach described in \cite{si2022grasp} for the grasp-stability prediction task is to optimize the friction coefficients by matching the grasping labels between simulated and real data under the same configuration of grasping heights and forces. We leave the exploration of better and more general ways for sim-to-real calibration as future work.


{\small
\bibliographystyle{ieee_fullname}
\bibliography{ref.bib}

\begin{thebibliography}{10}\itemsep=-1pt

\bibitem{arandjelovic2017look}
Relja Arandjelovic and Andrew Zisserman.
\newblock Look, listen and learn.
\newblock In {\em ICCV}, 2017.

\bibitem{dar}
Yusuf Aytar, Carl Vondrick, and Antonio Torralba.
\newblock See, hear, and read: Deep aligned representations.
\newblock {\em arXiv preprint arXiv:1706.00932}, 2017.

\bibitem{barbu2019objectnet}
Andrei Barbu, David Mayo, Julian Alverio, William Luo, Christopher Wang, Dan
  Gutfreund, Josh Tenenbaum, and Boris Katz.
\newblock {ObjectNet}: A large-scale bias-controlled dataset for pushing the
  limits of object recognition models.
\newblock In {\em NeurIPS}, 2019.

\bibitem{chamfer_distance}
Harry~G Barrow, Jay~M Tenenbaum, Robert~C Bolles, and Helen~C Wolf.
\newblock Parametric correspondence and {Chamfer} matching: Two new techniques
  for image matching.
\newblock In {\em Proceedings: Image Understanding Workshop}, pages 21--27,
  1977.

\bibitem{besl1992method}
Paul~J Besl and Neil~D McKay.
\newblock Method for registration of {3-D} shapes.
\newblock {\em IEEE Transactions on Pattern Analysis and Machine Intelligence},
  14:239--256, 1992.

\bibitem{calandra2018more}
Roberto Calandra, Andrew Owens, Dinesh Jayaraman, Justin Lin, Wenzhen Yuan,
  Jitendra Malik, Edward~H Adelson, and Sergey Levine.
\newblock More than a feeling: Learning to grasp and regrasp using vision and
  touch.
\newblock {\em RA-L}, 2018.

\bibitem{calandra2017feeling}
Roberto Calandra, Andrew Owens, Manu Upadhyaya, Wenzhen Yuan, Justin Lin,
  Edward~H Adelson, and Sergey Levine.
\newblock The feeling of success: Does touch sensing help predict grasp
  outcomes?
\newblock In {\em CoRL}, 2017.

\bibitem{chang2015shapenet}
Angel~X Chang, Thomas Funkhouser, Leonidas Guibas, Pat Hanrahan, Qixing Huang,
  Zimo Li, Silvio Savarese, Manolis Savva, Shuran Song, Hao Su, et~al.
\newblock {ShapeNet}: An information-rich {3D} model repository.
\newblock {\em arXiv preprint arXiv:1512.03012}, 2015.

\bibitem{chen2022visual}
Changan Chen, Ruohan Gao, Paul Calamia, and Kristen Grauman.
\newblock Visual acoustic matching.
\newblock In {\em CVPR}, 2022.

\bibitem{chen2020generating}
Peihao Chen, Yang Zhang, Mingkui Tan, Hongdong Xiao, Deng Huang, and Chuang
  Gan.
\newblock Generating visually aligned sound from videos.
\newblock {\em IEEE Transactions on Image Processing}, 2020.

\bibitem{chen2022sound}
Ziyang Chen, David~F Fouhey, and Andrew Owens.
\newblock Sound localization by self-supervised time delay estimation.
\newblock In {\em ECCV}, 2022.

\bibitem{choy20163d}
Christopher~B Choy, Danfei Xu, JunYoung Gwak, Kevin Chen, and Silvio Savarese.
\newblock {3D-R2N2}: A unified approach for single and multi-view {3D} object
  reconstruction.
\newblock In {\em ECCV}, 2016.

\bibitem{nuswide}
Tat-Seng Chua, Jinhui Tang, Richang Hong, Haojie Li, Zhiping Luo, and Yantao
  Zheng.
\newblock {NUS-WIDE}: A real-world web image database from national university
  of singapore.
\newblock In {\em Proceedings of the ACM International Conference on Image and
  Video Retrieval}, 2009.

\bibitem{cohen2003restricted}
David Cohen-Steiner and Jean-Marie Morvan.
\newblock Restricted delaunay triangulations and normal cycle.
\newblock In {\em Proceedings of the nineteenth annual symposium on
  Computational geometry}, 2003.

\bibitem{collins2021abo}
Jasmine Collins, Shubham Goel, Achleshwar Luthra, Leon Xu, Kenan Deng, Xi
  Zhang, Tomas~F Yago~Vicente, Himanshu Arora, Thomas Dideriksen, Matthieu
  Guillaumin, and Jitendra Malik.
\newblock {ABO}: Dataset and benchmarks for real-world {3D} object
  understanding.
\newblock {\em arXiv preprint arXiv:2110.06199}, 2021.

\bibitem{coumans2016pybullet}
Erwin Coumans and Yunfei Bai.
\newblock Pybullet, a python module for physics simulation for games, robotics
  and machine learning.
\newblock 2016.

\bibitem{de2005tutorial}
Pieter-Tjerk De~Boer, Dirk~P Kroese, Shie Mannor, and Reuven~Y Rubinstein.
\newblock A tutorial on the cross-entropy method.
\newblock {\em Annals of Operations Research}, 2005.

\bibitem{plsca}
Sijmen de Jong, Barry~M. Wise, and N.~Lawrence Ricker.
\newblock Canonical partial least squares and continuum power regression.
\newblock {\em Journal of Chemometrics: A Journal of the Chemometrics Society},
  2001.

\bibitem{deng2009imagenet}
Jia Deng, Wei Dong, Richard Socher, Li-Jia Li, Kai Li, and Li Fei-Fei.
\newblock {ImageNet}: A large-scale hierarchical image database.
\newblock In {\em CVPR}, 2009.

\bibitem{dong2017improved}
Siyuan Dong, Wenzhen Yuan, and Edward~H Adelson.
\newblock Improved gelsight tactile sensor for measuring geometry and slip.
\newblock In {\em IROS}, 2017.

\bibitem{downs2022google}
Laura Downs, Anthony Francis, Nate Koenig, Brandon Kinman, Ryan Hickman, Krista
  Reymann, Thomas~B McHugh, and Vincent Vanhoucke.
\newblock Google scanned objects: A high-quality dataset of {3D} scanned
  household items.
\newblock In {\em ICRA}, 2022.

\bibitem{ebert2018visual}
Frederik Ebert, Chelsea Finn, Sudeep Dasari, Annie Xie, Alex Lee, and Sergey
  Levine.
\newblock Visual foresight: Model-based deep reinforcement learning for
  vision-based robotic control.
\newblock {\em arXiv preprint arXiv:1812.00568}, 2018.

\bibitem{evans2022context}
Ben Evans, Abitha Thankaraj, and Lerrel Pinto.
\newblock Context is everything: Implicit identification for dynamics
  adaptation.
\newblock In {\em ICRA}, 2022.

\bibitem{finn2017deep}
Chelsea Finn and Sergey Levine.
\newblock Deep visual foresight for planning robot motion.
\newblock In {\em ICRA}, 2017.

\bibitem{gao2021ObjectFolder}
Ruohan Gao, Yen-Yu Chang, Shivani Mall, Li Fei-Fei, and Jiajun Wu.
\newblock {ObjectFolder}: A dataset of objects with implicit visual, auditory,
  and tactile representations.
\newblock In {\em CoRL}, 2021.

\bibitem{gao2018objectSounds}
Ruohan Gao, Rogerio Feris, and Kristen Grauman.
\newblock Learning to separate object sounds by watching unlabeled video.
\newblock In {\em ECCV}, 2018.

\bibitem{gao2019co}
Ruohan Gao and Kristen Grauman.
\newblock Co-separating sounds of visual objects.
\newblock In {\em ICCV}, 2019.

\bibitem{gao2022ObjectFolderV2}
Ruohan Gao, Zilin Si, Yen-Yu Chang, Samuel Clarke, Jeannette Bohg, Li Fei-Fei,
  Wenzhen Yuan, and Jiajun Wu.
\newblock {ObjectFolder} 2.0: A multisensory object dataset for sim2real
  transfer.
\newblock In {\em CVPR}, 2022.

\bibitem{girshick2014rich}
Ross Girshick, Jeff Donahue, Trevor Darrell, and Jitendra Malik.
\newblock Rich feature hierarchies for accurate object detection and semantic
  segmentation.
\newblock In {\em CVPR}, 2014.

\bibitem{greff2022kubric}
Klaus Greff, Francois Belletti, Lucas Beyer, Carl Doersch, Yilun Du, Daniel
  Duckworth, David~J Fleet, Dan Gnanapragasam, Florian Golemo, Charles
  Herrmann, et~al.
\newblock Kubric: A scalable dataset generator.
\newblock In {\em CVPR}, 2022.

\bibitem{he2017mask}
Kaiming He, Georgia Gkioxari, Piotr Doll{\'a}r, and Ross Girshick.
\newblock Mask {R-CNN}.
\newblock In {\em ICCV}, 2017.

\bibitem{he2016deep}
Kaiming He, Xiangyu Zhang, Shaoqing Ren, and Jian Sun.
\newblock Deep residual learning for image recognition.
\newblock In {\em CVPR}, 2016.

\bibitem{cca}
Harold Hotelling.
\newblock Relations between two sets of variates.
\newblock {\em Biometrika}, 28(3/4):321--377, 1936.

\bibitem{iashin2021taming}
Vladimir Iashin and Esa Rahtu.
\newblock Taming visually guided sound generation.
\newblock In {\em BMVC}, 2021.

\bibitem{pix2pix}
Phillip Isola, Jun-Yan Zhu, Tinghui Zhou, and Alexei~A Efros.
\newblock Image-to-image translation with conditional adversarial networks.
\newblock In {\em CVPR}, 2017.

\bibitem{jin2022neuralsounda}
Xutong Jin, Sheng Li, Guoping Wang, and Dinesh Manocha.
\newblock Neuralsound: Learning-based modal sound synthesis with acoustic
  transfer, May 2022.

\bibitem{krizhevsky2017imagenet}
Alex Krizhevsky, Ilya Sutskever, and Geoffrey~E Hinton.
\newblock Imagenet classification with deep convolutional neural networks.
\newblock {\em Communications of the ACM}, 2017.

\bibitem{kumar2019melgan}
Kundan Kumar, Rithesh Kumar, Thibault {de Boissiere}, Lucas Gestin, Wei~Zhen
  Teoh, Jose Sotelo, Alexandre {de Brebisson}, Yoshua Bengio, and Aaron
  Courville.
\newblock Melgan: Generative adversarial networks for conditional waveform
  synthesis, Dec. 2019.

\bibitem{kuznetsova2020open}
Alina Kuznetsova, Hassan Rom, Neil Alldrin, Jasper Uijlings, Ivan Krasin, Jordi
  Pont-Tuset, Shahab Kamali, Stefan Popov, Matteo Malloci, Alexander
  Kolesnikov, et~al.
\newblock The open images dataset v4.
\newblock {\em IJCV}, 2020.

\bibitem{lee2019touching}
Jet-Tsyn Lee, Danushka Bollegala, and Shan Luo.
\newblock “{Touching} to see” and “seeing to feel”: Robotic cross-modal
  sensory data generation for visual-tactile perception.
\newblock In {\em ICRA}, 2019.

\bibitem{lee2019making}
Michelle~A Lee, Yuke Zhu, Krishnan Srinivasan, Parth Shah, Silvio Savarese, Li
  Fei-Fei, Animesh Garg, and Jeannette Bohg.
\newblock Making sense of vision and touch: Self-supervised learning of
  multimodal representations for contact-rich tasks.
\newblock In {\em ICRA}, 2019.

\bibitem{li2022seehearfeel}
Hao Li, Yizhi Zhang, Junzhe Zhu, Shaoxiong Wang, A.~Michelle Lee, Huazhe Xu,
  Edward Adelson, Fei-Fei Li, Ruohan Gao, and Jiajun Wu.
\newblock See, hear, and feel: Smart sensory fusion for robotic manipulation.
\newblock In {\em CoRL}, 2022.

\bibitem{li2019connecting}
Yunzhu Li, Jun-Yan Zhu, Russ Tedrake, and Antonio Torralba.
\newblock Connecting touch and vision via cross-modal prediction.
\newblock In {\em CVPR}, 2019.

\bibitem{lin2014microsoft}
Tsung-Yi Lin, Michael Maire, Serge Belongie, James Hays, Pietro Perona, Deva
  Ramanan, Piotr Doll{\'a}r, and C~Lawrence Zitnick.
\newblock Microsoft {COCO}: Common objects in context.
\newblock In {\em ECCV}, 2014.

\bibitem{point_filtering}
Jun~S Liu and Rong Chen.
\newblock Sequential monte carlo methods for dynamic systems.
\newblock {\em Journal of the American Statistical Association},
  93(443):1032--1044, 1998.

\bibitem{luo2015localizing}
Shan Luo, Wenxuan Mou, Kaspar Althoefer, and Hongbin Liu.
\newblock Localizing the object contact through matching tactile features with
  visual map.
\newblock In {\em ICRA}, 2015.

\bibitem{manocha2021cdpam}
Pranay Manocha, Zeyu Jin, Richard Zhang, and Adam Finkelstein.
\newblock {CDPAM}: Contrastive learning for perceptual audio similarity.
\newblock In {\em ICASSP}, 2021.

\bibitem{mescheder2019occupancy}
Lars Mescheder, Michael Oechsle, Michael Niemeyer, Sebastian Nowozin, and
  Andreas Geiger.
\newblock Occupancy networks: Learning 3d reconstruction in function space.
\newblock In {\em CVPR}, 2019.

\bibitem{mildenhall2020nerf}
Ben Mildenhall, Pratul~P. Srinivasan, Matthew Tancik, Jonathan~T. Barron, Ravi
  Ramamoorthi, and Ren Ng.
\newblock {NeRF}: Representing scenes as neural radiance fields for view
  synthesis.
\newblock In {\em ECCV}, 2020.

\bibitem{neff2021donerf}
Thomas Neff, Pascal Stadlbauer, Mathias Parger, Andreas Kurz, Joerg~H Mueller,
  Chakravarty R~Alla Chaitanya, Anton Kaplanyan, and Markus Steinberger.
\newblock {DONeRF}: Towards real-time rendering of compact neural radiance
  fields using depth oracle networks.
\newblock In {\em EGSR}, 2021.

\bibitem{owens2018audio}
Andrew Owens and Alexei~A Efros.
\newblock Audio-visual scene analysis with self-supervised multisensory
  features.
\newblock In {\em ECCV}, 2018.

\bibitem{pai2001scanning}
Dinesh~K Pai, Kees van~den Doel, Doug~L James, Jochen Lang, John~E Lloyd,
  Joshua~L Richmond, and Som~H Yau.
\newblock Scanning physical interaction behavior of {3D} objects.
\newblock In {\em SIGGRAPH}, 2001.

\bibitem{park2019deepsdf}
Jeong~Joon Park, Peter Florence, Julian Straub, Richard Newcombe, and Steven
  Lovegrove.
\newblock {DeepSDF}: Learning continuous signed distance functions for shape
  representation.
\newblock In {\em CVPR}, 2019.

\bibitem{peng2017overview}
Yuxin Peng, Xin Huang, and Yunzhen Zhao.
\newblock An overview of cross-media retrieval: Concepts, methodologies,
  benchmarks, and challenges.
\newblock {\em TCSVT}, 2017.

\bibitem{peng2018modalityspecific}
Yuxin Peng, Jinwei Qi, and Yuxin Yuan.
\newblock Modality-specific cross-modal similarity measurement with recurrent
  attention network.
\newblock {\em TIP}, 2018.

\bibitem{Wikipedia}
Jose~Costa Pereira, Emanuele Coviello, Gabriel Doyle, Nikhil Rasiwasia, Gert~RG
  Lanckriet, Roger Levy, and Nuno Vasconcelos.
\newblock On the role of correlation and abstraction in cross-modal multimedia
  retrieval.
\newblock {\em IEEE TPAMI}, 36(03):521--535, 2014.

\bibitem{pascal_sentence}
Cyrus Rashtchian, Peter Young, Micah Hodosh, and Julia Hockenmaier.
\newblock Collecting image annotations using amazon's mechanical turk.
\newblock In {\em Proceedings of the NAACL HLT Workshops}, 2010.

\bibitem{reizenstein2021common}
Jeremy Reizenstein, Roman Shapovalov, Philipp Henzler, Luca Sbordone, Patrick
  Labatut, and David Novotny.
\newblock Common objects in {3D}: Large-scale learning and evaluation of
  real-life {3D} category reconstruction.
\newblock In {\em ICCV}, 2021.

\bibitem{rustler2022active}
Lukas Rustler, Jens Lundell, Jan~Kristof Behrens, Ville Kyrki, and Matej
  Hoffmann.
\newblock Active visuo-haptic object shape completion.
\newblock {\em RA-L}, 2022.

\bibitem{si2021taxim}
Zilin Si and Wenzhen Yuan.
\newblock Taxim: An example-based simulation model for gelsight tactile
  sensors.
\newblock {\em arXiv preprint arXiv:2109.04027}, 2021.

\bibitem{si2022grasp}
Zilin Si, Zirui Zhu, Arpit Agarwal, Stuart Anderson, and Wenzhen Yuan.
\newblock Grasp stability prediction with sim-to-real transfer from tactile
  sensing.
\newblock In {\em IROS}, 2022.

\bibitem{sitzmann2019scene}
Vincent Sitzmann, Michael Zollh{\"o}fer, and Gordon Wetzstein.
\newblock Scene representation networks: Continuous {3D}-structure-aware neural
  scene representations.
\newblock In {\em NeurIPS}, 2019.

\bibitem{smith2021active}
Edward Smith, David Meger, Luis Pineda, Roberto Calandra, Jitendra Malik,
  Adriana Romero~Soriano, and Michal Drozdzal.
\newblock Active {3D} shape reconstruction from vision and touch.
\newblock {\em NeurIPS}, 2021.

\bibitem{smith20203d}
Edward~J Smith, Roberto Calandra, Adriana Romero, Georgia Gkioxari, David
  Meger, Jitendra Malik, and Michal Drozdzal.
\newblock {3D} shape reconstruction from vision and touch.
\newblock In {\em NeurIPS}, 2020.

\bibitem{smith2005development}
Linda Smith and Michael Gasser.
\newblock The development of embodied cognition: Six lessons from babies.
\newblock {\em Artificial Life}, 2005.

\bibitem{pix3d}
Xingyuan Sun, Jiajun Wu, Xiuming Zhang, Zhoutong Zhang, Chengkai Zhang, Tianfan
  Xue, Joshua~B Tenenbaum, and William~T Freeman.
\newblock {Pix3D}: Dataset and methods for single-image {3D} shape modeling.
\newblock In {\em CVPR}, 2018.

\bibitem{suresh2021efficient}
Sudharshan Suresh, Zilin Si, Joshua~G Mangelson, Wenzhen Yuan, and Michael
  Kaess.
\newblock Efficient shape mapping through dense touch and vision.
\newblock {\em arXiv preprint arXiv:2109.09884}, 2021.

\bibitem{Suresh22icra}
Sudharshan Suresh, Zilin Si, Joshua~G Mangelson, Wenzhen Yuan, and Michael
  Kaess.
\newblock {ShapeMap 3-D}: Efficient shape mapping through dense touch and
  vision.
\newblock In {\em ICRA}, 2022.

\bibitem{tian2019manipulation}
Stephen Tian, Frederik Ebert, Dinesh Jayaraman, Mayur Mudigonda, Chelsea Finn,
  Roberto Calandra, and Sergey Levine.
\newblock Manipulation by feel: Touch-based control with deep predictive
  models.
\newblock In {\em ICRA}, 2019.

\bibitem{tian2020contrastive}
Yonglong Tian, Dilip Krishnan, and Phillip Isola.
\newblock Contrastive multiview coding.
\newblock In {\em ECCV}, 2020.

\bibitem{villegas2019high}
Ruben Villegas, Arkanath Pathak, Harini Kannan, Dumitru Erhan, Quoc~V Le, and
  Honglak Lee.
\newblock High fidelity video prediction with large stochastic recurrent neural
  networks.
\newblock {\em NeurIPS}, 2019.

\bibitem{wang2020tacto}
Shaoxiong Wang, Mike Lambeta, Po-Wei Chou, and Roberto Calandra.
\newblock {TACTO}: A fast, flexible and open-source simulator for
  high-resolution vision-based tactile sensors.
\newblock {\em arXiv preprint arXiv:2012.08456}, 2020.

\bibitem{grady2016mppi}
Grady Williams, Paul Drews, Brian Goldfain, James~M. Rehg, and Evangelos~A.
  Theodorou.
\newblock Aggressive driving with model predictive path integral control.
\newblock In {\em ICRA}, 2016.

\bibitem{wu20153d}
Zhirong Wu, Shuran Song, Aditya Khosla, Fisher Yu, Linguang Zhang, Xiaoou Tang,
  and Jianxiong Xiao.
\newblock {3D} {ShapeNets}: A deep representation for volumetric shapes.
\newblock In {\em CVPR}, 2015.

\bibitem{fenet}
Yong Xu, Feng Li, Zhile Chen, Jinxiu Liang, and Yuhui Quan.
\newblock Encoding spatial distribution of convolutional features for texture
  representation.
\newblock In {\em NeurIPS}, 2021.

\bibitem{yang2022touch}
Fengyu Yang, Chenyang Ma, Jiacheng Zhang, Jing Zhu, Wenzhen Yuan, and Andrew
  Owens.
\newblock Touch and go: Learning from human-collected vision and touch.
\newblock In {\em NeurIPS Datasets and Benchmarks Track}, 2022.

\bibitem{yuan2017gelsight}
Wenzhen Yuan, Siyuan Dong, and Edward~H Adelson.
\newblock Gelsight: High-resolution robot tactile sensors for estimating
  geometry and force.
\newblock {\em Sensors}, 2017.

\bibitem{pcn}
Wentao Yuan, Tejas Khot, David Held, Christoph Mertz, and Martial Hebert.
\newblock {PCN}: Point completion network.
\newblock In {\em 3DV}, 2018.

\bibitem{zhang2017shape}
Zhoutong Zhang, Qiujia Li, Zhengjia Huang, Jiajun Wu, Joshua~B Tenenbaum, and
  William~T Freeman.
\newblock Shape and material from sound.
\newblock In {\em NeurIPS}, 2017.

\bibitem{zhao2018sound}
Hang Zhao, Chuang Gan, Andrew Rouditchenko, Carl Vondrick, Josh McDermott, and
  Antonio Torralba.
\newblock The sound of pixels.
\newblock In {\em ECCV}, 2018.

\bibitem{dscmr}
Liangli Zhen, Peng Hu, Xu Wang, and Dezhong Peng.
\newblock Deep supervised cross-modal retrieval.
\newblock In {\em CVPR}, 2019.

\bibitem{zhou2018visual}
Yipin Zhou, Zhaowen Wang, Chen Fang, Trung Bui, and Tamara~L Berg.
\newblock Visual to sound: Generating natural sound for videos in the wild.
\newblock In {\em CVPR}, 2018.

\end{thebibliography}
}

\end{document}